\documentclass[sigconf, 10pt]{acmart}
\AtBeginDocument{%
  \providecommand\BibTeX{{%
    \normalfont B\kern-0.5em{\scshape i\kern-0.25em b}\kern-0.8em\TeX}}}

\setcopyright{none}
\renewcommand\footnotetextcopyrightpermission[1]{© Guilherme H. Apostolo 2025. This is the author's version of the work. It is posted here for your personal use. Not for redistribution. The definitive Version of Record was published in the 31st Annual International Conference on Mobile Computing and Networking (ACM MOBICOM '25), November 4--8, 2025, Hong Kong, China, \url{http://dx.doi.org/10.1145/3680207.3765260}.} 

\acmConference[MobiCom'25]{31st Annual International Conference on Mobile Computing and Networking}{Nov 2025}{Hong Kong, China}
\usepackage{multirow}
\usepackage[normalem]{ulem}
\usepackage{array}
\usepackage{tabularx}
\usepackage{xcolor}
\usepackage[algoruled,vlined,linesnumbered, boxed, noend, algo2e]{algorithm2e}

\acmSubmissionID{208}

\usepackage{subcaption}
\usepackage[export]{adjustbox}
\usepackage{tipa}
\usepackage{graphicx}
\usepackage{soul}
\usepackage{xcolor}
\usepackage{tcolorbox}
\usepackage{textcomp}
\usepackage{tipa}
\usepackage{oplotsymbl}
\usepackage{tikz}
\usepackage{enumitem}

\newcommand{\name}{Uirapuru}

\newcommand\refAlgorithm[1]{Alg.~\ref{alg:#1}}

\newcommand\refFigure[1]{Fig.~\ref{fig:#1}}
\newcommand\refSection[1]{\S\ref{sec:#1}}
\newcommand\refSubsection[1]{\S\ref{subsec:#1}}
\newcommand\refTable[1]{Tab.~\ref{tab:#1}}

\ifdefined\commentsenabled
\newcommand\todo[1]{\textcolor{red}{TODO: #1}}
\newcommand\guilherme[1]{\textcolor{blue}{ #1}}
\newcommand\lin[1]{\textcolor{green}{Lin: #1}}
\newcommand\pablo[1]{\textcolor{purple}{Pablo: #1}}
\newcommand\vinod[1]{\textcolor{orange}{Vinod: #1}}
\else
\newcommand\todo[1]{}
\newcommand\guilherme[1]{\textcolor{blue}{}}
\newcommand\lin[1]{}
\newcommand\pablo[1]{}
\newcommand\vinod[1]{}
\fi

\newcommand{\network}{M}

\newcommand{\meannetworklatency}{L^{\text{mean}}_{M}}
\newcommand{\percentilenetworklatency}{L^{\text{99}}_{M}}
\newcommand{\networkaccuracy}{A_{M}}
\newcommand{\accuracybincount}{b}
\newcommand{\accuracybin}[1]{\alpha_{S_#1}}

\newcommand{\globalview}{V_{\text{global}}}
\newcommand{\localview}{V_{\text{frame}}}
\newcommand{\transformation}{T}
\newcommand{\transformationdelta}{\Delta T}
\newcommand{\localtransformation}{T_{\text{frame}}}

\newcommand{\localobjects}{0}
\newcommand{\tileview}{W}
\newcommand{\objectdistribution}{D_{\localobjects}^{\tileview}}
\newcommand{\tileestimatedaccuracy}{\textit{eAP}^{\tileview}}
\newcommand{\plan}{\tileview_{p}}
\newcommand{\planestimatedaccuracy}{PeAP}
\newcommand{\objectpercentage}{\lambda_{\tileview}}

\newcommand{\aliasconservative}{\textbf{UC}}
\newcommand{\aliasnonconservative}{\textbf{UNC}}
\newcommand{\aliasdownsample}{\textbf{DS}}
\newcommand{\aliasuniform}{\textbf{UT}}
\newcommand{\aliasremix}{\textbf{R}}

\newcommand{\myparagraph}[1]{\textbf{#1.}}
\newcommand{\ballnumber}[1]{\tikz[baseline=(myanchor.base)] \node[circle,fill=.,inner sep=0.5pt] (myanchor) {\color{white}\bfseries\footnotesize #1};}

\hypersetup{pdfstartview=FitH,
  pdfpagelayout=SinglePage,
  bookmarksnumbered=true,
  bookmarksopen=true,
  colorlinks=true
}

\newlength{\onecolgrid}
\newlength{\twocolgrid}
\newlength{\threecolgrid}
\newlength{\fourcolgrid}
\begin{document}

\title{\name{}: Timely Video Analytics for High-Resolution Steerable Cameras on Edge Devices}


\author{Guilherme H. Apostolo}
    \affiliation{%
    \institution{Vrije Universiteit Amsterdam}
    \country{}}
    \email{g.apostolo@vu.nl}

\author{Pablo Bauszat}
    \affiliation{%
    \institution{Vrije Universiteit Amsterdam}
    \country{}}
    \email{pablo.bauszat@gmail.com}

\author{Vinod Nigade}
    \affiliation{%
    \institution{Vrije Universiteit Amsterdam}
    \country{}}
    \email{vinod.nigade@gmail.com}

\author{Henri E. Bal}
    \affiliation{%
    \institution{Vrije Universiteit Amsterdam}
    \country{}}
    \email{h.e.bal@vu.nl}

\author{Lin Wang}
    \affiliation{%
    \institution{Paderborn University}
    \country{}}
    \email{lin.wang@uni-paderborn.de}



\begin{abstract}

Real-time video analytics on high-resolution cameras has become a popular technology for various intelligent services like traffic control and crowd monitoring. While extensive work has been done on improving analytics accuracy with timing guarantees, virtually all of them target static viewpoint cameras. In this paper, we present \name{}, a novel framework for real-time, edge-based video analytics on high-resolution \emph{steerable} cameras. The actuation performed by those cameras brings significant dynamism to the scene, presenting a critical challenge to existing popular approaches such as frame tiling. To address this problem, \name{} incorporates a comprehensive understanding of camera actuation into the system design paired with fast adaptive tiling at a per-frame level. We evaluate \name{} on a high-resolution video dataset, augmented by pan-tilt-zoom (PTZ) movements typical for steerable cameras and on real-world videos collected from an actual PTZ camera. Our experimental results show that \name{} provides up to $1.45\times$ improvement in accuracy while respecting specified latency budgets or reaches up to $4.53\times$ inference speedup with on-par accuracy compared to state-of-the-art static camera approaches.

\end{abstract}




\maketitle


\section{Introduction}
\label{sec:introduction}

Recent technological advances have introduced high-reso\-lu\-tion cameras (e.g., 4K, 8K, and higher) that are rapidly spreading around public places like traffic intersections, airports, and nursing homes~\cite{2018-sigcomm-chameleon,2019-pieee-safety,2019-cvpr-power-of-tilling,2023-infocom-cross-camera-edge}. 
Public and private organizations utilize these cameras for tasks like traffic control, crowd monitoring, and security and safety~\cite{2017-nsdi-jetstream,2021-mmsys-cross-roi,2020-sec-spatula}. The high-resolution video streams generated by these cameras are typically processed by machine learning models such as deep neural networks, performing video analytics tasks including object detection, tracking, and activity recognition~\cite{2016-cvpr-resnet,2019-cvpr-tracking,2017-cvpr-action}. 
These applications typically impose strict real-time requirements on video analytics, a.k.a. service-level objectives (SLOs), calling for a camera-local solution given the high variability of the communication network for cloud-based alternatives~\cite{2021-mmsys-cross-roi, 2021-mobicom-remix,2018-icdcs-efficient-tiling,2022-eurosys-litereconfig,2023-infocom-cross-camera-edge}. Security and data privacy concerns further reinforce this need~\cite{2020-arxiv-amadeus}.

\begin{figure}[!t]
    \centering
    \includegraphics[width=0.45\textwidth,page=1]{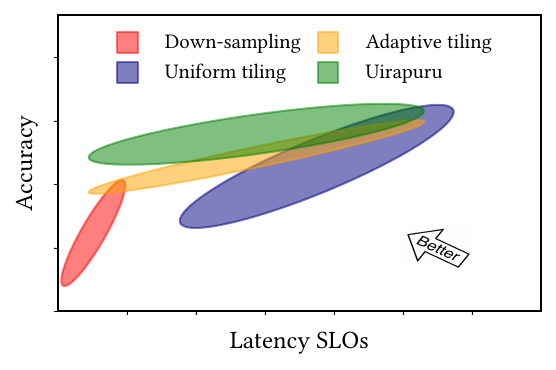}
    \caption{A comparison of \name{} to previous approaches in terms of accuracy and latency SLOs.}
    \Description{A comparison of \name{} to previous approaches in terms of accuracy and latency SLOs.}
    \label{fig:tiling-issue}
\end{figure}

Meeting the stringent timing requirements with on-device processing poses a significant challenge due to the mismatch between the required intensive computation and the limited available resources~\cite{2022-infocom-flexpatch}. 
One straightforward approach for dealing with this issue is to use small and fast models and down-sample the video frames. However, down-sampling will most likely degrade the quality of the analytics due to the loss of details, thus defeating the purpose of high-resolution cameras. 
A more involved approach is \emph{tiling}, where a high-resolution video frame is broken down into smaller tiles which are processed individually with smaller models~\cite{2018-hpec-4k-tiling,2019-cvpr-power-of-tilling, 2021-mobicom-remix, 2018-icdcs-efficient-tiling}.
Adaptive tiling exploits the \emph{spatial and temporal locality} in the object distribution of a scene and often provides improved accuracy over simple uniform tiling. 
By customizing the tile layout (e.g., using fine-grained tiles only for crowded regions) and applying scene-specific optimizations (e.g., skipping tiles that do not contain objects of interest), video analytics can be accelerated significantly. 

Many of the cameras deployed today are \emph{steerable}~\cite{2023-acm-actrack,2017-ipsn-panotes,2011-mmsys-steerable}.
Steerable cameras can perform human-instructed actuation such as pan-tilt-zoom (PTZ)~\cite{2020-cvpr-deepptz,2010-signal-processing-ptz}. 
The actuation performed by a steerable camera present significant challenges to tiling-based video analytics due to the high dynamism that is introduced.
More specifically, the object distribution in the video stream of a steerable camera is constantly changing, making it hard to leverage spatial and temporal locality.

We propose \textbf{\name{}}\footnote{\name{} (\textipa{/wi.Ra\textprimstress pu.Ru/}) or Musician Wren is a native bird of the Amazon rainforest. The bird has a complex and varied song, with locals claiming it can sing for minutes without repeating a single note in its melody (just like our system that can create a new plan for every frame).}, a novel framework for real-time video analytics on high-resolution steerable cameras. 
\name{} addresses the aforementioned challenges to provide high-performance video analytics for steerable cameras while guaranteeing latency SLOs.
To account for camera actuation, \name{} first introduces the concepts of a ``global'' view perspective that encompasses the whole scene of observation and a ``local'' view perspective that reflects the current view of the camera. 
\name{} builds the global view with historical frames and later, at runtime, transforms the objects from the global view to the local view according to the camera's actuation. 
This allows \name{} to estimate the object distribution of each frame more accurately in real-time.  
Based on these distributions, \name{} dynamically generates a tile plan per frame that assigns more accurate models to regions of interest, using a novel fast tiling algorithm based on dynamic programming.
\name{} also features a new model profile that better encompasses the high variability of object sizes introduced by camera actuation. 
As shown in Fig.~\ref{fig:tiling-issue}, this novel design allows \name{} to surpass previous state-of-the-art solutions in video analytics scenarios with high-resolution steerable cameras.

Overall, we make the following contributions:
\begin{itemize}[leftmargin=*]
    %
    %
    \item We present \name{}, a framework for edge-based, latency-critical video analytics for high-resolution steerable cameras that incorporates an understanding of camera actuation and addresses profiling biases in its design (\S\ref{sec:system}). 
    \item We propose a fast and optimal tiling algorithm, named \textbf{d}epth-first \textbf{p}ost-order \textbf{d}ynamic \textbf{p}rogramming (DP-DP), that efficiently generates high-quality tile plans (\S\ref{subsec:tile-planner}).
    \item We demonstrate the effectiveness of our prototype by conducting extensive experiments with a high-resolution object detection dataset retrofitted with actuation by steerable cameras and on a real-world steerable camera deployment~(\S\ref{sec:evaluation}). Our results show an improvement in accuracy up to 1.4x while meeting user-specified latency SLOs, and reach an up to $4.53\times$ inference speedup with on-par accuracy compared to the baseline.
\end{itemize}


\section{Background and Related Work}
\label{sec:background}

Deep learning has become the de-facto standard for many video analytics tasks. 
To address the computational challenge of deep learning models, existing works have explored offloading model inference to a cloud platform~\cite{2020-sensys-distream, 2021-mobicom-elf,2020-sec-clownfish,2017-sec-parkmaster,2023-ieee-edgeduet,2016-mob-glimpse,2020-sec-spatula,2019-sensys-caesar}. 
Yet, such offloading-based approaches face major challenges centered around SLO guarantee (due to network variability in the wide area) and data privacy. 
In this paper, we focus on video analytics solely on edge devices, specifically on high-resolution steerable cameras.

\subsection{Edge-based Video Analytics}

Performing analytics with deep learning on the original full-resolution video data (in 4K, 8K, or even higher) on edge devices is prohibitively expensive and impractical, considering the limited onboard resources~\cite{2022-eurosys-litereconfig,2022-infocom-flexpatch,2021-mobicom-remix,2019-cvpr-power-of-tilling}. 
Popular approaches in the literature that address this challenge are \emph{down-sampling} and \emph{tiling}.

Given a model tailored for a target edge device, down-sampling reduces the frame resolution to match the model's input size. 
This provides opportunities for latency guarantees, albeit at the cost of significant accuracy losses due to drastic reductions in object sizes in the input. 
To improve accuracy, one must increase the model size or use more sophisticated architectures~\cite{2016-cvpr-resnet,2017-ieee-fasterr-cnn,2020-cvpr-efficientdet,2019-icml-efficientnet}, provided that latency requirements of the application and memory constraints of the edge device are still met. 
However, the capacities of the edge device impose strict limitations that make it generally hard to scale up the models.
For example, popular mobile system-on-chips such as Snapdragon 855~\cite{2024-website-snapdragon-855} and Kirin 970~\cite{2024-website-kirin-970} can run the EfficientDet family~\cite{2020-cvpr-efficientdet} only from D0 up to D5~\cite{2021-mobicom-remix}.
At the same time, the Nvidia Jetson AGX~\cite{2024-website-jetson-AGX} can run up to D7 but with a high latency of over 1.5 seconds per frame.

\begin{figure*}[!t]
    \begin{minipage}[t]{0.6\textwidth}
        \centering
        \includegraphics[width=\textwidth]{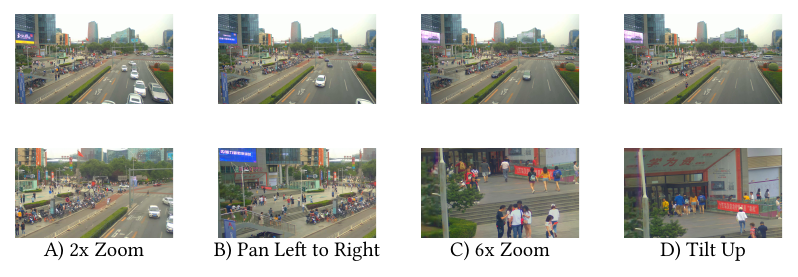}  
    \end{minipage}
    \begin{minipage}[t]{0.35\textwidth}
        \centering
        \includegraphics[width=\textwidth]{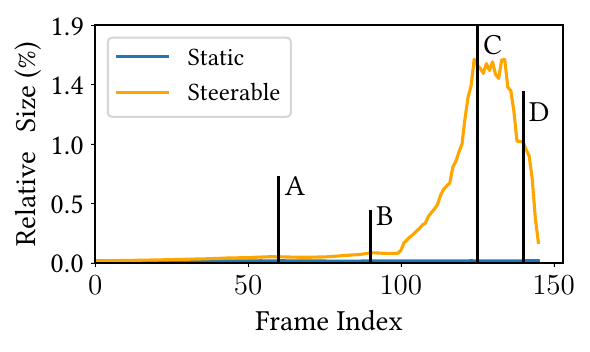} 
    \end{minipage}
    \caption{The figure shows video frames from a static camera (top) and a steerable camera executing a sequence of actuation (bottom) on the left side. The right plot shows the average size of objects in each frame of the sequence.}
    \Description{The figure shows video frames from a static camera (top) and a steerable camera executing a sequence of actuation (bottom) on the left side. The right plot shows the average size of objects in each frame of the sequence.}
    \label{fig:objects_size_motivation}
\end{figure*}

Alternatively, tiling has been utilized to perform high-resolution video analytics on the edge.
Tiling breaks down the frame into smaller tiles, allowing the use of smaller models based on the tile size~\cite{2018-hpec-4k-tiling,2019-cvpr-power-of-tilling}. 
Existing tiling solutions can be roughly categorized into \emph{uniform} and \emph{adaptive}.
Uniform tiling generates equally-sized tiles whose resolution roughly matches the expected input size of a specific model.
Despite achieving higher accuracy, uniform tiling suffers from high latency since the per-frame inference time is the sum of all tile inference times.
Moreover, uniform tiling fails to provide a fine-grained trade-off between accuracy and latency.
To meet latency SLOs between the execution time of two models of different sizes, we are forced to choose the smaller model with degraded accuracy since using the larger model would cause SLO violations.

Adaptive tiling approaches~\cite{2018-icdcs-efficient-tiling,2021-cvpr-dmnet,2023-ieee-edgeduet} allow tiles of varying sizes and shapes, as well as tile-specific model selection.
For example, tiles with higher importance can be allocated more of the time budget and computing resources, and be processed with more sophisticated models to improve accuracy. 
However, such flexibility comes at the cost of higher complexity in the tile generation. 
Remix is a prominent example of adaptive tiling~\cite{2021-mobicom-remix} for high-resolution static videos and guides its tile plan generation with historical information collected from video frames during a bootstrap phase.
While achieving superior accuracy compared to down-sampling and uniform tiling-based approaches, Remix assumes strong temporal correlations in the video stream for its optimizations to work, which may not hold in dynamic scenarios. 
%


\subsection{Challenges with Steerable Cameras}
\label{sec:motivation}

Steerable cameras that support actuation (i.e., sequence of movements) like pan-tilt-zoom (PTZ)~\cite{2023-acm-actrack,2020-cvpr-deepptz} have been increasingly deployed. 
Such cameras enable essential use cases, such as close inspection of specific regions in the camera view by applying user-instructed actuation, but bring additional challenges to high-resolution video analytics due to the significantly increased dynamism.

Existing solutions based on adaptive tiling typically leverage spatial and temporal locality in video frames for tile generation. 
However, videos from steerable cameras may lack such a locality. 
\refFigure{objects_size_motivation} depicts the frames from a static camera (top row) and a steerable camera executing a sequence of actuation (bottom row).
On the right side of the figure, the plot shows how the average object size (relative to the frame resolution) changes in every frame.
For the static camera case, the average object size and density of objects in the scene remain almost constant.
Previous approaches, such as Remix, leverage such spatial and temporal locality and rely on a stable object distribution when generating tiles.
In contrast, frames captured by steerable cameras have diverging view points due to the actuation applied to the camera and exhibit drastic changes in object sizes and locations as seen in \refFigure{objects_size_motivation}.
\emph{This renders existing approaches for tile generation based on historical object distributions impractical}. 

When comparing the object distribution of a frame to the historical objects collected from a global viewpoint during an early bootstrap phase, changes typically occur in two ways: (1) changing object density in particular regions of the scene and (2) changing object size. 
We can see examples of the first case in \refFigure{objects_size_motivation}(B) with a pan movement from left to right and \refFigure{objects_size_motivation}(D) with a tilt movement to the top. 
For the second case, we can observe changes in object sizes in \refFigure{objects_size_motivation}(A, C) caused by a zoom movement and in \refFigure{objects_size_motivation}(D) due to a tilt movement. 
In extreme cases, an object can become even bigger than the tile containing it, likely leading to the model's complete failure.
Existing adaptive tiling approaches assume relatively static object distributions, so tiles that are anticipated to contain few objects are processed with less accurate models, and conversely, tiles that are assumed to cover densely populated areas are processed with highly accurate models. 
Mismatches in the anticipated number of objects and their sizes can lead to accuracy degradation (when sparsely populated areas become crowded with essential objects) and resource waste (when previously crowded regions become empty).
\textit{This shows that tile planning cannot purely rely on historical object information in the case of steerable cameras.} 

Moreover, the rapid changes and high range of object sizes bring extra challenges to the performance profiling of models~\cite{2022-rtss-jellyfish, 2021-mobicom-remix,2018-sec-video-edge}. 
Previous approaches do already distinguish model performance by object size, but typically categorize objects into uniform ranges using absolute pixel values.
For example, Remix~\cite{2021-mobicom-remix} uses a profile with 12 uniform accuracy bins, the smallest being between $0-64$ pixels and the biggest between $196^{2}-\infty$.
When applying models to tiles that are only a subset of the image (where objects are now relatively enlarged), they propose using a correcting scaling factor to address the difference in resolution and view.
Their results demonstrate that such a profile is effective enough for the static camera scenario where object sizes are roughly constant and bounded.
However, in the steerable camera scenario, sizes vary significantly in terms of absolute pixel values and over a much larger range.
For example, an object that was far away before and minuscule can be a quarter of the image's area size after a zoom actuation.
Additionally, there are cases where zooming can cause objects to become even larger than the model inputs themselves. 
In such cases, tiling can be detrimental, and down-sampling would be a better solution.
Such variations have the effect that objects are quickly projected to belong to the largest bin, eliminating any ability for distinction when comparing models.
\textit{Therefore, it is integral to create model profiles that are well suited for rapid, highly variable size changes and understand how changes in the relative sizes of objects affect model accuracy.}
%


\section{\name{} System Design}
\label{sec:system}

\begin{figure*}[!t]
    \centering
    \includegraphics[page=3, width=0.8\textwidth]{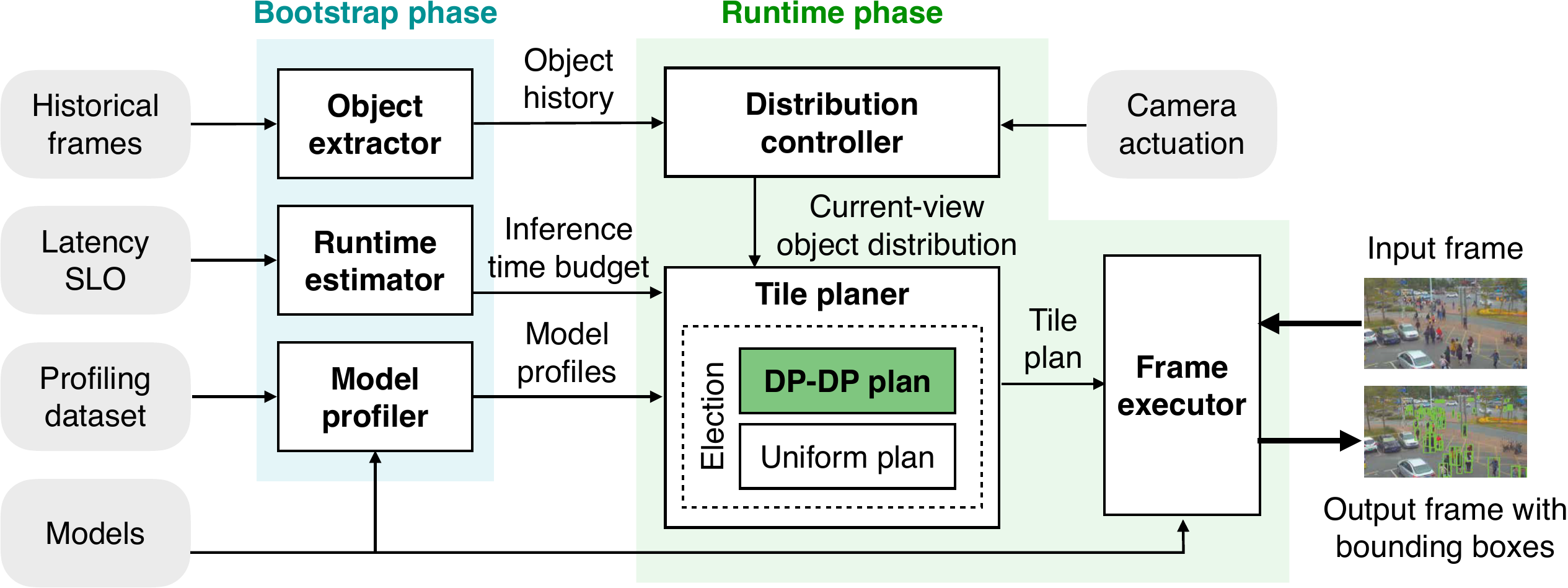}
    \caption{\name{} system overview.}
    \Description{\name{} system overview.}
    \label{fig:systems-overview}
\end{figure*}

The main goal of our \textbf{\name{}} system is to maximize object detection accuracy for steerable cameras within the available resources of an edge device given a per-frame latency SLO.
To achieve this, \name{} must deal with rapidly changing object distributions (i.e., locations and sizes) arising from camera actuation.
Our key idea is to keep track of these actuation over time in order to understand the camera's view transformation in each frame.
Initially, \name{} collects a historical object distribution from a set of frames, denoted as the historical frames, that views the scene from a global, all-encompassing perspective.
In each frame, we then extract and transform the relevant historical objects utilizing the state of the camera to form a \emph{local} object distribution (``local'' here contrasts a frame's view in relation to the ``global'' view perspective).
The local distribution identifies regions of interest much more timely and accurately and is then utilized for adaptive tiling.

Previous approaches that perform adaptive tiling create tile plans during an offline phase, typically of very high quality ~\cite{2021-mobicom-remix}.
However, these plans do not adapt well when cameras are steered.
We argue that a timely plan is more beneficial than an optimal one as long as it still provides a sufficient level of quality.
Accordingly, we propose an approximate but fast \emph{online} adaptive tiling strategy that efficiently generates a fresh tile layout for each frame.

Our system's overall design and components are illustrated in \refFigure{systems-overview}.
\name{} operates in two phases: an offline bootstrap phase~(\refSection{bootstrap}) and a runtime phase~(\refSection{runtime-phase}).
The former is done as a pre-process to generate model profiles and extract objects from historical frames for a global overview.
The latter processes the incoming camera frames through tiling and performs the actual model inference.

\subsection{Bootstrap Phase}
\label{sec:bootstrap}

During the bootstrap phase, \name{} first creates a performance profile for each model in the provided family of models using a profiling dataset (\refSubsection{network-profiler}). 
Each profile contains the model's characteristics in terms of inference latency and detection accuracy for different object sizes. It is later used during the tile plan creation to decide on the most suitable models.
Afterward, the objects in the historical frames are extracted (\refSubsection{historical-frames}).
These objects provide a global overview of where and in which size objects are to be expected. They will be used later, at runtime, to estimate the local object distribution in each frame.
Finally, as generating a new tile plan in each frame consumes a part of the available latency budget, we also use the historical frames to estimate the algorithmic runtime overhead of our plan creation method (\refSubsection{plan-runtime-estimation}).
This guarantees that during the runtime phase, the system understands the available \emph{inference} budget that already accounts for the overhead from the plan creation algorithm and does not violate the per-frame SLO.
The inputs to the bootstrap phase are: 1) a family of models for the object detection task, 2) a profiling dataset, 3) the historical frames taken from the global view perspective, and 4) the per-frame latency SLO.

\subsubsection{Model Profiler}
\label{subsec:network-profiler}

For each model $\network$ in the family, we create a model profile that stores the models’ mean and $99$\textit{th} percentile inference latencies, $\meannetworklatency$, and $\percentilenetworklatency$, and an accuracy vector with $\accuracybincount$ bins for different size ranges, $\networkaccuracy$:
\begin{align*}
    \networkaccuracy &= \langle \accuracybin{0}, \accuracybin{1}, \cdots, \accuracybin{{b-1}} \rangle
\end{align*}

\begin{figure}[ht]
    \centering
    \includegraphics[width=\twocolgrid,page=1]{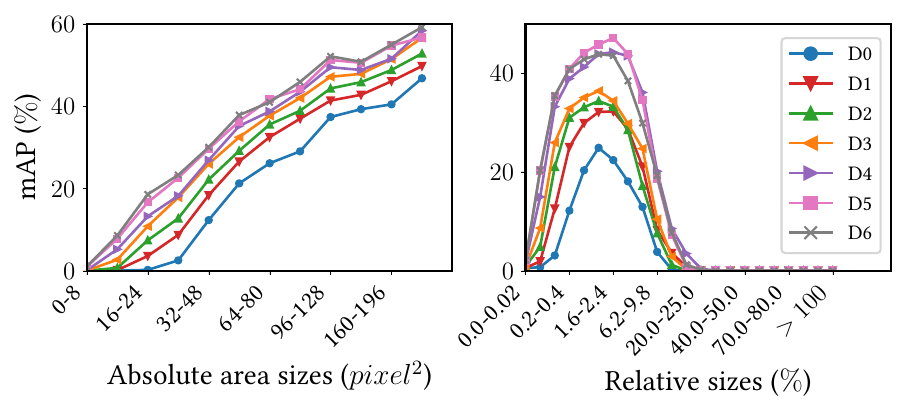}
    \caption{Accuracy profiles using Remix's absolute pixel sizes (left) vs. \name{}'s relative sizes (right).}
    \Description{Accuracy profiles using Remix's absolute pixel sizes (left) vs. \name{}'s relative sizes (right).}
    \label{fig:profiles-comparison}
\end{figure}

The inference latencies are measured by collecting the mean and $99$\textit{th} percentile of the inference time across multiple executions of the models on the edge device.
We will only use one of them during runtime depending on the working mode of \name{} that defines the strictness of the SLO guarantee that the system should provide (see \refSubsection{tile-planner}).
The accuracy vector $\networkaccuracy$ represents the model's estimated accuracy for various ranges of object sizes.
Models tend to perform differently depending on the size of objects, e.g., small models are generally good at detecting relatively large objects while tiny objects are often missed \cite{2019-cvpr-power-of-tilling}.
Hence, it is beneficial to estimate the mean accuracy of a model over all objects, which allows for more fine-grained distinction based on object size.

In the steerable camera scenario, sizes vary significantly in terms of absolute pixel values and are over a much larger range.
Such variations have the effect that objects are quickly projected to belong to the largest bin, eliminating any ability for distinction when comparing models.
To tackle those issues, we propose using a non-linear, relative size model profile to address both problems and better represent model performance for wide-ranging object sizes.
Using relative sizes, our model profile is easily transferable between different views and tile sizes without the need for a correcting scaling factor.
Further, using a non-linear profile allows fine-grained distinction for small sizes, which is required for down-sampling, while still covering a wide range of sizes, as is required for tiling.
Our profile consists of 22 bins.
The profile can use any non-linear function to define its bins.
For ease of comparison with the other baseline in~\refSubsection{ablation-profile},
the ranges for the first 12 bins are computed by taking the 12 absolute value ranges from Remix divided by the resolution of Efficientdet-D6, while bins 13 and 14 ranges from $20\%-25\%$ and $25\%-30\%$ respectively.
The following eight bins are chosen from a linear range of $10\%$ until $>100\%$.
\refFigure{profiles-comparison} shows Remix and \name{} profiles over the training data. 
\refFigure{profiles-comparison} shows that the Remix profile captures only one side of the profile curve, limiting its ability to evaluate models when actuation and tiling make objects too large.
For instance, an object with an absolute area size of 600 on our profile will be profiled into $\alpha$ at $21.9\%$ on Efficientdet-D6, while on Efficientdet-D0, it exceeds $\alpha > 100\%$.

Our model profiler generates sub-images for each input image from the profiling dataset, ensuring that the models see various content and sizes.
The collection used to profile each model consists of original images and sub-images, created by uniformly partitioning the original ones for each available model resolution.

\subsubsection{Historical Frames Object Extraction}
\label{subsec:historical-frames}

\name{} needs to identify regions of interest and the sizes of objects in these regions in each frame to create a tile plan.
We assume that the object distribution of the historical frames, provided or collected during the bootstrap phase, is a good overall representative of how objects will appear in the scene.
Further, we assume that, during these historical frames, the camera transformation is in an \emph{identity} state.
This means that the historical frames are either viewed from a global perspective, representing the widest angle of view that encapsulates all the scenes of interest, or a collection of frames with smaller angles that can represent the global perspective when combined.
The collection of frames with smaller angles does not need to be captured simultaneously but rather in a short period.
The historical frames need to represent the distribution of objects in the scene correctly.
Therefore, as long as the capture of those frames is done in a short period of time, this does not pose challenges to \name{} performance as demonstrated later in~\refSubsection{ablation-historical-frames}.
As the camera deployer typically controls the placement and understands the scene, we believe this is a reasonable assumption for this to be feasible.

\name{} collects all objects from the historical frames during the bootstrap phase.
The best way to extract those objects initially would be to use an Oracle model; however, as this is not possible for real-case scenarios, we use the best strategy available to us.
In our case, we employ the uniform partition strategy using the most accurate model possible.
Previous approaches demonstrate that such choice, while approximate, is enough to provide reliable results \cite{2021-mobicom-remix,2021-mobicom-elf}.

\subsubsection{Plan Runtime Estimation}
\label{subsec:plan-runtime-estimation}

\name{} is designed to generate plans rapidly so every frame is executed with a new tile plan.
Theoretically, one would have to update the tile plan only when a camera actuation happens. 
In practice, however, the camera is often constantly in motion (e.g., panning or tilting slowly across a region of interest) so we chose to re-compute a new plan each frame.
Although our algorithm is fast, it still introduces a non-negligible overhead that we need to subtract from the total per-frame latency SLO to get to the actual budget for inference.
As the plan creation might vary depending on the number of objects in a frame, we use the historical frames to estimate the introduced overhead.
We first collect the time to execute the tiling algorithm in each historical frame using the latency SLO.
Then, we use the $99$\textit{th} percentile of the times, plus $10\%$ of its value as a safety tolerance, as the cost for plan creation.

\subsection{Runtime Phase}
\label{sec:runtime-phase}

\name{}'s runtime phase generates a tile plan, runs inference on the tiles, and merges the detection results for every incoming frame.
First, camera transformations triggered by actuation are tracked using the distribution controller, and the current camera state is used to extract the historical objects relevant to the current frame's view (\refSubsection{distribution-controller}).
Using these objects, the tile planer generates a fresh tiling layout that maximizes accuracy while respecting the available latency budget (\refSubsection{tile-planner}).
\name{} has two working modes that depend on the chosen SLO guarantee: a more stringent, conservative mode (for a per-frame latency guarantee) and a relaxed, non-conservative mode (for an average inference time guarantee).
Finally, each tile is inferred with its associated model, and the individual detections are merged to produce the final results (\refSubsection{inference}).
The inputs to the runtime phase are: 1) a family of models with its corresponding profiles, 2) the historical global view objects, 3) the current camera frame with its corresponding actuation, 4) the latency budget, and 5) the desired SLO working mode.

\subsubsection{Distribution Controller}
\label{subsec:distribution-controller}

The object distributor controller has two functions: 1) transform the objects extracted from the historical frames in the global view coordinates into local coordinates, and 2) generate the tile object distribution for each tile position requested by the tile planner. 
As the camera operator applies an actuation, the objects in the view will change in size and appearance.
Therefore, the system adapts historical objects by applying the same transformations, ensuring consistency between global and local distributions.
This design allows \name{} to execute the transformation only for objects detected in historical frames and not for objects detected by the camera during the runtime phase, which avoids problems with discontinuity or fast-moving objects such as cars. 
For example, fast-moving cars will be captured over several historical frames, their locations and sizes identified in the global view distribution, and then later transformed to the local distribution if the street appears in it.
As long as \name{} collects enough historical frames, the distribution controller will be able to identify the distributions of objects in all camera regions correctly, providing reliable input for adaptive tiling.

The distribution controller tracks the camera transformation $\transformation$ over time.
We assume that the camera initially starts in a rest position, i.e., $\transformation$ is a 3D identity transformation that corresponds to the global scene perspective with a camera view $\globalview$. 
Each actuation by the camera operator causes a camera movement, potentially extending over multiple frames, that introduces a change to the transformation $\transformationdelta$.
This change $\transformationdelta$ is passed as additional input to the controller together with each image frame.
The goal is then to extract the relevant objects from the global historical frames and transform their coordinates into the local coordinates of the current frame.

Since we do not have full access to the 3D coordinates of the historical objects, but only their 2D bounding boxes, we can only approximately estimate their new 2D locations using re-projection~\cite{2022-urban-3d-recons,2004-cambridge-view-geometry}.
The distribution controller computes the local 3D transformation of a frame, $\localtransformation$, by tracking and accumulating $\transformationdelta$.
Assuming a simple pinhole camera model, we compute the local view of a frame, $\localview$, by applying $\localtransformation$ to the original view of the camera $\globalview$.
Afterward, we re-project the 2D coordinates of the bounding boxes of the historical objects from the plane of $\globalview$ onto the local plane defined by $\localview$.
Objects that lie outside of $\localview$ are removed as they have no impact on the tile planning.
While our approach is based on an approximate estimation and may be potentially affected by issues such as object occlusions and distortions, it achieves a good trade-off between computational efficiency and practical accuracy in our scenario.
We acknowledge the use of more sophisticated computer graphics and vision techniques (e.g., motion estimation~\cite{2011-ieee-visual-odometry}, view synthesis~\cite{2014-cvpr-view-synthesis}, image-based rendering~\cite{2000-spie-image-rendering}, or deep learning~\cite{2020-cvpr-deepptz}) as interesting avenues for future work.

\subsubsection{Tile Planer}
\label{subsec:tile-planner}

The tile planner slices the high-resolution frame into a collection of non-overlapping image regions (tiles) and selects a model for each tile in such a way that it achieves the highest possible (estimated) accuracy while guaranteeing execution within the required latency.
Exploring all the possible tile layouts and model combinations is a time-consuming task that can take up to minutes~\cite{2021-mobicom-remix}.
One reason for that is that previous approaches typically link the arrangement of tiles directly with the model selection, i.e., tiles are sized according to the available models.
Here, our main insight is that, for fast performance, we have to \textbf{separate the tile arrangement from the model selection} to reduce the search space of combinations significantly.
We propose to first perform the division of the image independently using some form of hierarchical space partitioning, disregarding the available model sizes.
This reduces the possible combinations of tile arrangements while allowing for a reasonable amount of space exploration and adaptation.
Afterward, model selection is performed, but no longer influences the chosen regions.
Instead, tiles are fit to models through resizing (up- or down-sampling).
While resizing can introduce some accuracy degradation, these effects are implicitly integrated into the model selection with our new profile.
\pablo{sentence could be clearer.}
Even with a reduced number of possible layouts, however, brute-force searching for the best combination of tiles and models still requires an infeasible amount of time.
Therefore, we propose a recursive dynamic programming solution to solve the problem more efficiently.

\myparagraph{Hierarchical Space Partitioning}
First, we chose the quad-tree as a simple way of hierarchically partitioning the image into regions~\cite{1984-acm-quadtree}, but other hierarchical tree structures would also work.
The quad-tree is a tree structure where each node has four children that subdivide the space of the parent into four non-overlapping quadrants.
Each leaf node in a quad-tree corresponds to a tile region that needs to be inferred.
We select a maximum tree depth based on the resolution of the input frame (in our case we use a depth of 3 for 4K input images), and explore all possible topologies as options for tile layouts.

\myparagraph{Tile Plan}
For each node, we can compute its distribution of object sizes by first finding the objects falling into the node's region $\tileview$ from the set of local objects $\localobjects$, and then define a vector $\objectdistribution$ with $\accuracybincount{b}$ elements, where each element represents the normalized number of objects ($\phi$) for each of the pre-defined sizes:
\begin{equation*}
    \objectdistribution = \langle \phi S_0, \phi S_1, \cdots, \phi S_{b-1} \rangle
\end{equation*}
The object distribution of a node can then be combined with a model profile to generate a regionally-adaptive estimate of that model's accuracy:
\begin{equation*}
    \tileestimatedaccuracy = \objectdistribution * \networkaccuracy
\end{equation*}
The choice of tree topology and model selection in each leaf node constitute a tile plan.
The estimated accuracy of a plan, $\planestimatedaccuracy$, is simply the sum of all the tile's estimated accuracy weighted by the fraction of objects $\objectpercentage$ in each tile:
\begin{equation}
    \label{eq:-plan-estimated-accuracy}
    \planestimatedaccuracy = \sum_{\tileview \in \plan} \tileestimatedaccuracy \cdot \objectpercentage
\end{equation}

\myparagraph{DP-DP Algorithm} 
To optimally select a set of nodes (tiles) from a quad-tree topology and assign models to those nodes, we observe that this combinatorial problem can be formulated as a 0-1 knapsack problem~\cite{2004-spriner-knapsack} with additional constraints. 
In this context, the items in the ``knapsack'' are tuples consisting of nodes and models, where the weight of the item (tuple) is the model's inference latency, and the value of the item is the model's accuracy for that corresponding tile. 
The total inference latency budget represents the capacity of the knapsack. 
There are two additional constraints: \textbf{(C1)} a node cannot be selected if any of its descendant nodes are chosen, as this would create overlapping areas and lead to redundant execution; \textbf{(C2)} only one model can be selected for a node, so two tuples with the same tile cannot be chosen.
To solve this problem optimally in (pseudo-)polynomial time, we propose a dynamic programming-based (DP) solution that traverses the items (nodes) and builds the DP solution in a specific order to avoid constraint violation. 
We refer to this algorithm as the \textbf{d}epth-first \textbf{p}ost-order \textbf{d}ynamic \textbf{p}rogramming (DP-DP) algorithm.

The algorithm first determines each node's optimal assignment to find the model that maximizes its accuracy.
The algorithm then uses a depth-first post-order recursion to build the partial and optimal DP solution at each node.
This traversal order ensures that optimal solutions for child nodes are available at parent nodes and that partial optimal solutions for the left sub-tree or left sibling are available when we reach the right sub-tree nodes.
The algorithm ultimately returns an array with the optimal mapping of nodes to models, where every node is either assigned a model or marked as empty.
Appendix~\ref{appendix:pseudo} presents the recursive pseudo-code for the DP-DP algorithm.

\begin{figure}[!t]
    \centering
    \includegraphics[page=1, width=0.47\textwidth]{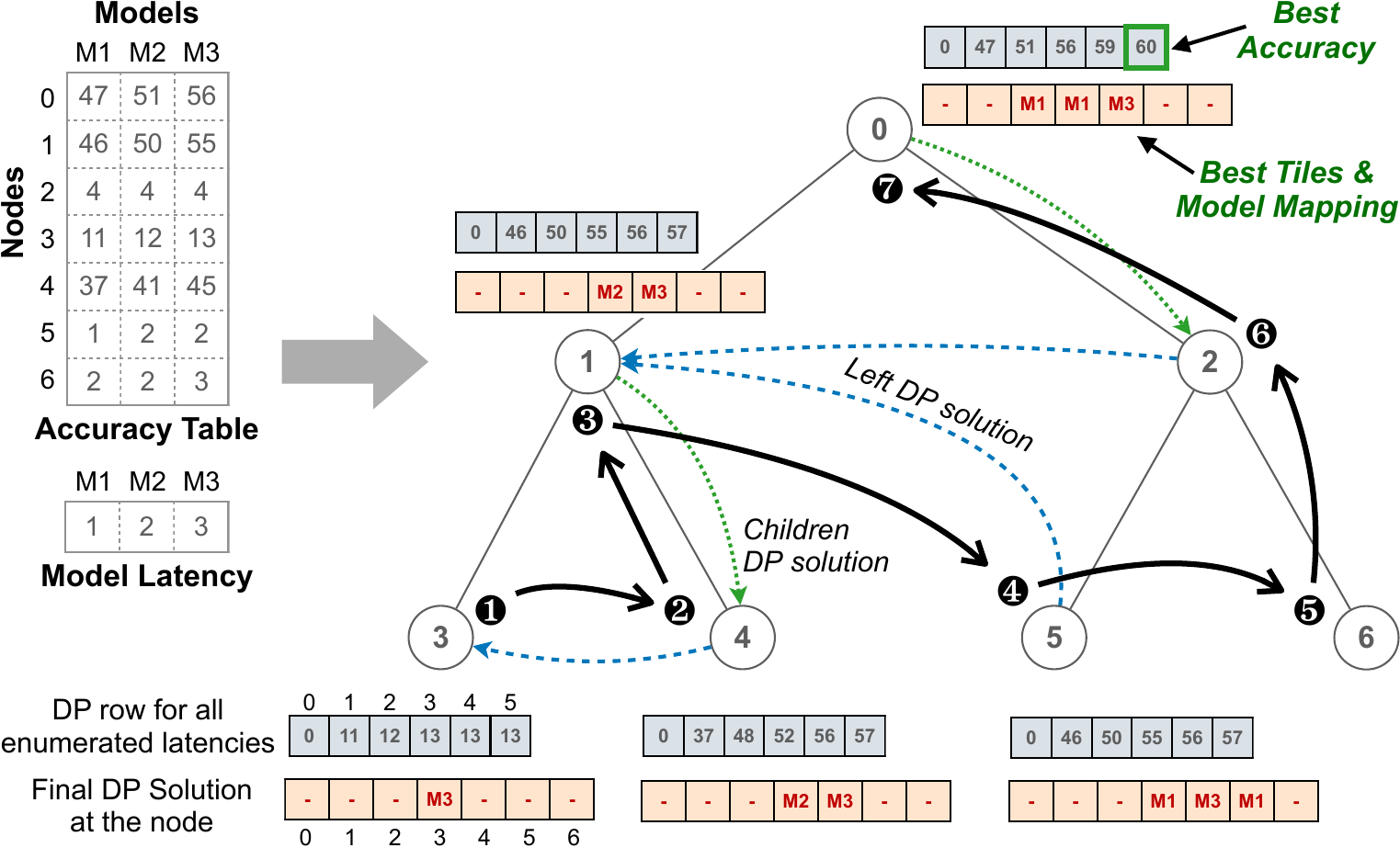}
    \caption{Illustration of the \textbf{d}epth-first \textbf{p}ost-order \textbf{d}ynamic \textbf{p}rogramming algorithm for a binary tree of depth 3.}
    \Description{Illustration of the \textbf{d}epth-first \textbf{p}ost-order \textbf{d}ynamic \textbf{p}rogramming algorithm for a binary tree of depth 3.}
    \label{fig:dp-dp:illustration}
\end{figure}

\myparagraph{Example}
\refFigure{dp-dp:illustration} illustrates a simple example of the DP-DP algorithm applied to a binary tree with three levels. 
The accuracy table contains the estimated accuracy for $7$ nodes and $3$ models. 
Note that the accuracy values differ across tiles (nodes), even for the same model, due to variations in object distribution and sizes. 
The latency vector indicates the inference latency of the three models. 
Our DP-DP algorithm starts \ballnumber{1} by finding the DP solution at Node~$3$. 
At Node~$3$, with a maximum latency budget of $5$, the algorithm selects Model M$1$, which has an inference latency of $3$ and accuracy of $13$. 
Moving \ballnumber{2} to the second node in the depth-first post-order traversal, Node~$4$, the algorithm selects the best combination of both the tiles (Node~$3$ and $4$) and chooses Model M$2$ and M$3$, respectively, resulting in a total latency of $5$ and aggregate accuracy of $57$. 
At the parent node $1$, the algorithm selects \ballnumber{3} the best solution between Nodes $1$ and $4$, retaining the solution from Node~$4$. 
At Node~$5$, the algorithm selects \ballnumber{4} the optimal combination of the solution at Node~$4$ and the best model (if any) at Node~$5$. 
Finally, following the recursive process to the root node, the algorithm determines \ballnumber{7} the optimal node-to-model mapping: Node~$2$ with Model~M$1$, Node $3$ with Model~$M1$, and Node~$4$ with Model~M$3$, achieving the highest accuracy of $60$.

In the worst case, the latency interval is $1$ unit when discretizing the latency values.
Therefore, the asymptotic complexity of our algorithm becomes $O( \mathbin\#Tiles \cdot |\mathbb{M}| \cdot L_{slo})$, where $L_{slo}$ is the inference latency budget.
\pablo{Should be node count, not tile count?}
For proof of the optimality of the DP-DP algorithm, we refer to the supplemental material found in Appendix~\ref{appendix:proof}.
\pablo{Add link to the supplemental material.}
\pablo{Unify the notation with the rest of the paper!}

\myparagraph{Working Modes}
The latency of a plan is simply estimated by summing up the inference times for all its tiles.
\pablo{We could mention batching here.}
However, \name{} has two working modes, non-conservative and conservative, that depends on which latency estimate from the model profile is chosen.
\name{}'s non-conservative mode uses the mean inference time of the model profile, $\meannetworklatency$.
While this allows plans to generally select larger models, it also makes tile plans more prone to underestimate a frame's real execution time. 
Ultimately, the system will converge towards an average execution time over the execution of a sequence of frames, but a per-frame latency is not guaranteed.
\name{}'s conservative mode guarantees that a certain percentile of frames fulfills the specified latency SLO by using $\percentilenetworklatency$, the more strict estimate of model inference time.
This forces the plans to select models more conservatively and makes them less prone to underestimate the real execution time.
By default, \name{}'s conservative profile uses the $99$\textit{th} percentile inference time estimate, and so the system should estimate the time spent on inference accurately for at least $99\%$ of frames, but other values could be chosen depending on the required SLO.

\myparagraph{Plan Election} 
\name{} additionally creates uniform plans that simply slice the high-resolution image into non-overlapping tiles for each model in the family.
The final tile plan for a frame is the most accurate between the adaptive and the best uniform plan.

\subsubsection{Frame Inference}
\label{subsec:inference}

Once the tile plan is established, the actual model inference can be performed.
The frame execution module slices the high-resolution image into tiles according to the plan layout and runs inference on each tile using the designated model. 
Padding is added before a tile is passed to a model to reduce the chance of missing objects at tile boundaries.
Padding is a common practice, and there are many ways to decide the size of the padding \cite{2021-mobicom-remix,2019-cvpr-power-of-tilling,2023-ieee-edgeduet}.
We chose a simple uniform padding, which provided sufficient results with minimum overhead.
The individual detection results from the tiles are merged using Non-Maximum Suppression~\cite{2015-ieee-fatsr-cnn} to avoid object duplicates.


\section{Evaluation}
\label{sec:evaluation}

We first evaluate \name{}'s performance in terms of real-time inference latency guarantees and accuracy and compare it to existing baselines.
We then perform an ablation study to examine the individual components' performance.
Furthermore, we conduct a case study using \name{} in a real-world PTZ camera.
Finally, we conducted an overhead analysis. 
We then perform an ablation study to examine the individual components' performance.

\subsection{Experimental Setup}

\begin{figure*}[!t]
    \centering
    \begin{subfigure}[b]{0.49\textwidth} 
        \centering
        \includegraphics[width=1\linewidth]{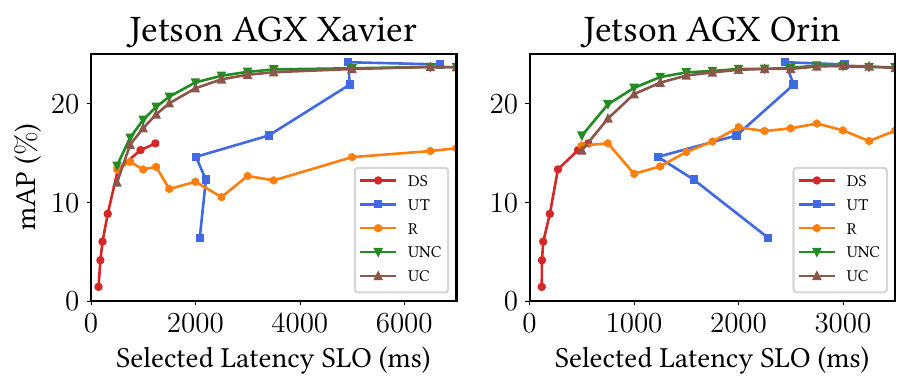} 
        \caption{\name{} vs. baselines}
        \label{fig:eval-overall}
    \end{subfigure}
    \hfill
    \begin{subfigure}[b]{0.49\textwidth}  
        \centering
        \includegraphics[width=1\linewidth]{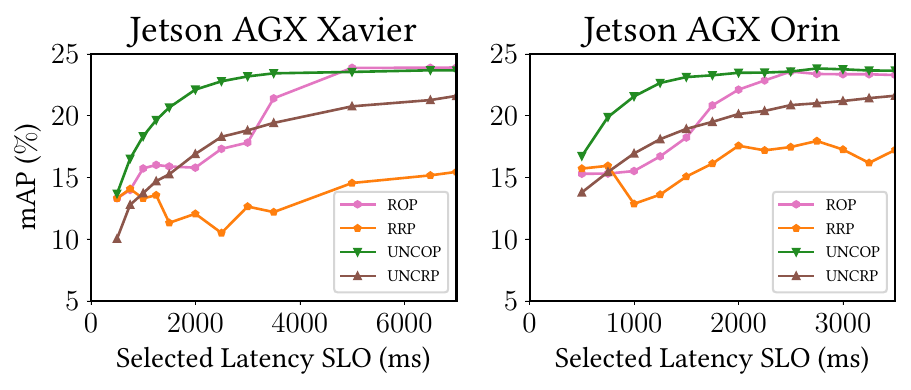}  
        \caption{\name{} vs. Remix using different model profiles}
        \label{fig:eval-profile}
    \end{subfigure}
    \caption{mAP performance of \name{} on the steerable PANDA dataset across different hardware.}
    \Description{mAP performance of \name{} on the steerable PANDA dataset with different hardware.}
    \label{fig:Overall-Profile}
\end{figure*}

\myparagraph{Hardware and Implementation} We evaluate our approach using two series of edge device hardware: the Jetson AGX Xavier~\cite{2024-website-jetson-AGX} and the Jetson AGX Orin~\cite{2024-website-jetson-orin} series. 
Jetson AGX Xavier has a 512-core NVIDIA Volta architecture GPU with 64 Tensor Cores GPU and an 8-core NVIDIA Carmel Arm®v8.2 64-bit CPU. In contrast, the Jetson AGX Orin has a 2048-core NVIDIA Ampere architecture GPU with 64 Tensor Cores and a 12-core Arm Cortex 64-bit CPU. 
Our \name{} system prototype is implemented using Python 3.8.
The models are built on top of Google's TensorFlow~\cite{2024-website-tensorflow} framework, and we use OpenCV~\cite{2024-website-opencv} for image processing.

\myparagraph{Dataset and Models}
We use the EfficientDet~\cite{2020-cvpr-efficientdet} family as our models for inference to be consistent with the Remix baseline. 
We use these variants ranging from D0 to D6 for all experiments.
Each variant has different input and memory sizes (see \refTable{efficientdet}).
The models are fine-tuned using the training and test images from the PANDA Image dataset~\cite{2020-cvpr-panda} using $600$ epochs.

\begin{table}[t]
    \centering
    \caption{Parameters of EfficientDet models.}
    \resizebox{\columnwidth}{!}{
        \begin{tabular}{@{}ccccccccc@{}}
            \toprule
            \textbf{Variant} & \textbf{D0} & \textbf{D1} & \textbf{D2} & \textbf{D3} & \textbf{D4} & \textbf{D5} & \textbf{D6} \\ \midrule
            \textbf{Input Size} & $512^2$ & $640^2$ & $768^2$ & $896^2$ & $1024^2$ & $1280^2$ & $1280^2$ \\
            \textbf{Size (MB)} & 61.3 & 92.2 & 112 & 154 & 235 & 359 & 503 \\
            \bottomrule
        \end{tabular}
    }
    \label{tab:efficientdet}
\end{table}

While existing datasets in the literature have some of the specific requirements of our scenario~\cite{2020-vision-uav-dataset,2021-cvpr-visdrone,2020-cvpr-panda}, none could fulfill all of them: high-resolution, steerable videos with a high density of objects, and pre-defined/controlled movements.
As a result of such limitation, we created a steerable video dataset based on the PANDA dataset~\cite{2020-cvpr-panda} for our evaluation, exploiting the fact that we have an extremely high image resolution available (see example in Appendix~\ref{appendix:dataset_figures}).
The original resolution of the video is $26,753$x$15,052$ pixels, allowing us to create smaller 4K cut-out videos that virtually represent transformations (i.e., pan, tilt, and zoom movements) that a steerable camera would apply.
In total, the PANDA dataset has $10$ video sequences.
We generate a random PTZ sequence for each of the $10$ sequences, where each randomly generated sequence contains at least one of the three types of actuation and, in some cases, more than one. 
The type and parameters of each actuation (i.e., direction, magnitude, start frame, and frame count) are randomly selected, but image boundaries are respected.
The duration of each actuation is randomly selected from a range of 15 and a maximum of 50 frames.
There is a 15-frame pause (no movement) between the actuation.

\myparagraph{Baselines}
We compare \name{} against the results obtained by three baseline approaches: down-sampling (\aliasdownsample{}), uniform tiling (\aliasuniform{}), and Remix (\aliasremix{}).

\myparagraph{Evaluation Parameters}
We evaluate 12 latency SLOs for Xavier: 500, 750, 1000, 1250, 1500, 2000, 2500, 3000, 3500, 5000, 6500, and 7000 ms, and 13 SLOs for Orin: 500, 750, 1000, 1250, 1500, 1750, 2000, 2250, 2500, 2750, 3000, 3250, and 3500 ms, covering from the smallest inference time with down-sampling to the largest uniform tiling for both hardware.
We used the mean averaged precision (mAP) and the average precision at IoU=0.50 (AP50), which are widely used metrics to evaluate video analysis tasks in industry~\cite{2024-website-coco}.
\subsection{Overall Performance}

We first evaluate the overall performance of \name{}'s non-conservative (\aliasnonconservative{}) and conservative (\aliasconservative{}) mode in terms of accuracy in \refFigure{eval-overall}.
Each point represents the achieved mAP for a latency SLO on the steerable PANDA dataset.
Here, we only report mAP as it is the most stringent of our metrics.
\name{}'s non-conservative version has an mAP improvement factor between $0.97$-$1.45\times$ compared to the closest baseline for both Xavier and Orin, outperforming especially in the SLO range from 500ms until 3500ms, with the highest improvement of $1.45\times$ being achieved at the 3000ms SLO on the Xavier.
For the Orin, \aliasnonconservative{} has a better mAP than the baselines in the range from 500ms to 2250ms, with the most significant improvement for an SLO of 1500ms.
\name{}'s conservative version performs very similarly, closely following the trend of its non-conservative counterpart.
The slight performance decrease is explained by the fact that \name{}'s conservative version has more stringent constraints on the models that can be selected. 
As the latency SLO grows, both versions have more options available, bringing the performance of \name{}'s conservative version closer to the non-conservative one.
Above the 5000ms SLO for the Xavier and 2500ms for the Orin, \name{} performance roughly converges towards the most expensive and accurate choice, which is uniform tiling (\aliasuniform{}) with EfficientDet-D5 and D6.

Overall, \name{} achieves higher mAP in a shorter time than the baselines.
\name{} is up to $4.53\times$ faster in terms of latency on the Jetson AGX Xavier hardware in reaching a certain mAP value compared to similar mAP values from the baselines within 1\% of tolerance.
For example, for the Xavier, the \aliasnonconservative{} mode achieves $16.49\%$ mAP at 750ms, while uniform tiling using EfficientDet-D3 achieves $16.71\%$ mAP requiring 3404ms.
Similarly, \aliasnonconservative{} reaches up to $2.53\times$ inference speedup while achieving on-par mAP for the Orin.

\subsubsection{Average Latency vs Per-frame SLO}
Remix and \name{}'s non-conservative mode both provide an average latency execution convergence, i.e., the average execution time per frame will ultimately become close to the latency SLO.
However, this means that occasionally, the inference time of a frame might exceed the specified SLO.
\name{}'s conservative mode provides a more strict per-frame latency guarantee.
In \refTable{slos}, we show the miss rate (i.e., the percentage of frames with a higher execution time than the allowed SLO) in percentage values for Remix (\aliasremix{}) and both our modes.
It can be seen that \aliasnonconservative{} delivers more than 99\% of the frames within the desired latency SLO.
\name{}'s non-conservative miss rate can be explained by its profile that takes the average inference time for models into account instead of the more stringent $99th$ percentile.
While \name{}'s non-conservative cannot provide such stringent guarantees, it still provides significantly lower miss rates than Remix.

\begin{table}[!t]
\centering
\caption{Miss rate (\%) for Remix (\aliasremix{}), \name{} Non-Conservative (\aliasnonconservative{}), and \name{} Conservative (\aliasconservative{}).}
\Description{Miss Rate (\%) for Remix (\aliasremix{}), \name{} Non-Conservative (\aliasnonconservative{}), and \name{} Conservative (\aliasconservative{})}
\small
\begin{tabular}{|c|ccc|ccc|}
\hline
\multirow{2}{*}{
\textbf{SLO}} & \multicolumn{3}{c|}{\textbf{AGX Xavier}} & \multicolumn{3}{c|}{\textbf{AGX Orin}} \\ \cline{2-7} 
    & \textbf{\aliasremix{}} & \textbf{\aliasnonconservative{}} & \textbf{\aliasconservative{}} & \textbf{\aliasremix{}} & \textbf{\aliasnonconservative{}} & \textbf{\aliasconservative{}} \\ \hline
500  & 0.61  & 7.04 & 0.49 & 64.48 & 3.90 & 0       \\ 
750  & 40.51 & 6.56 & 0.14 & 0     & 1.05 & 0.27    \\ 
1000 & 36.95 & 5.16 & 0    & 16.46 & 4.60 & 0.34    \\ 
1250 & 0     & 1.19 & 0.07 & 21.72 & 0.35 & 0       \\ 
1500 & 2.73  & 2.23 & 0.07 & 27.59 & 0.28 & 0.06    \\ 
1750 & -     & -    & -    & 25.68 & 0.07 & 0       \\ 
2000 & 8.33  & 2.65 & 0.14 & 23.49 & 0.21 & 0.06    \\ 
2250 & -     & -    & -    & 23.29 & 0    & 0       \\ 
2500 & 22.68 & 0.14 & 0.07 & 27.80 & 0    & 0.13    \\ 
2750 & -     & -    & -    & 26.02 & 0    & 0.13    \\ 
3000 & 20.97 & 0.49 & 0    & 25.88 & 0    & 0       \\ 
3250 & -     & -    & -    & 21.58 & 0.21 & 0       \\ 
3500 & 18.65 & 0.14 & 0    & 24.65 & 0    & 0.13    \\ 
5000 & 8.40  & 0    & 0    & -     & -    & -       \\ 
6500 & 13.73 & 0    & 0    & -     & -    & -       \\ 
7000 & 0     & 0.07 & 0.07 & -     & -    & -       \\ \hline
\end{tabular}
\label{tab:slos}
\end{table}

\subsubsection{PTZ Breakdown}
\label{subsec:ptz_breakdown}

\begin{figure}[t]
    \centering
    \includegraphics[width=\twocolgrid,page=1]{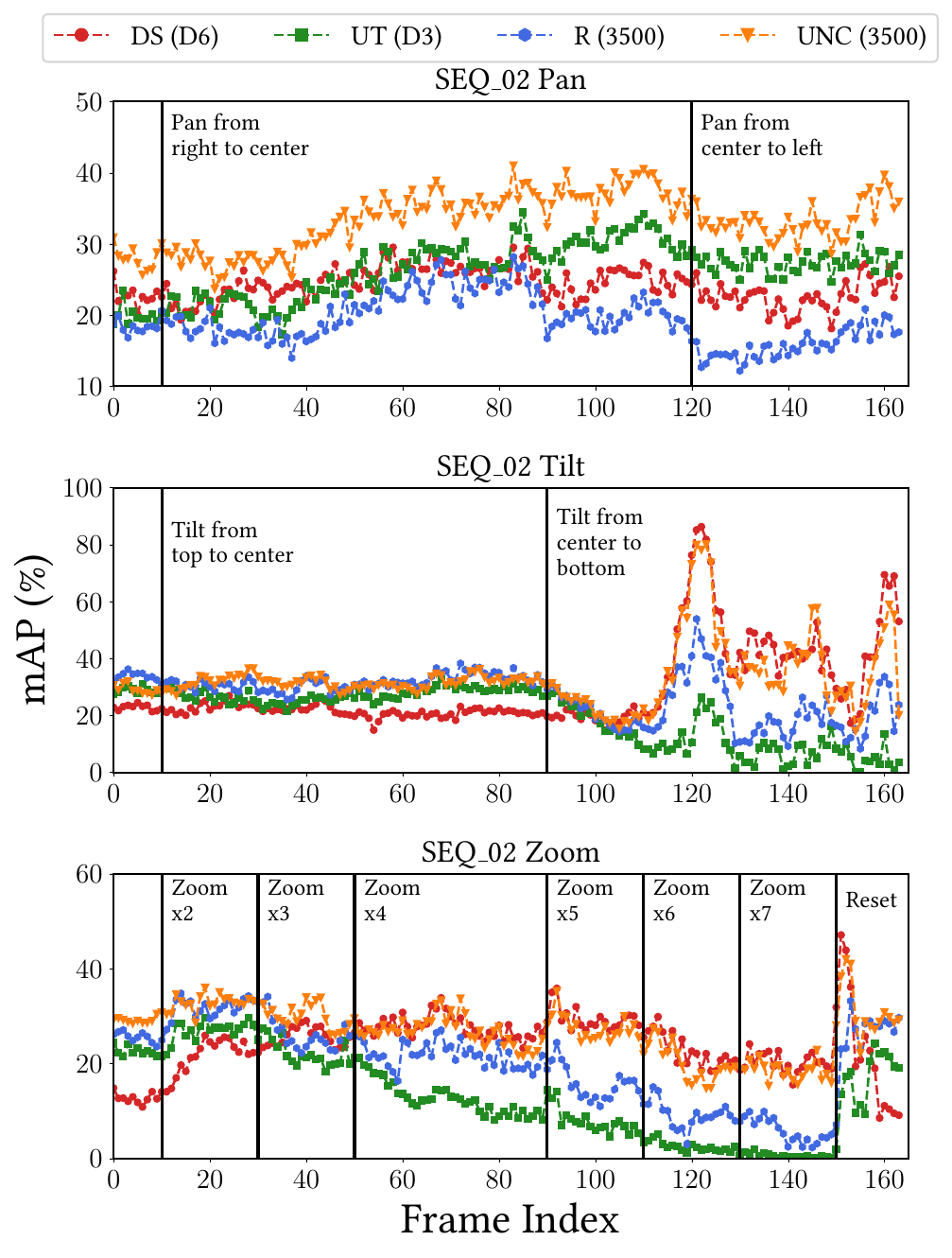}
    \caption{mAP performance per frame of \name{} for individual actuations on the Jetson AGX Xavier.}
    \Description{mAP performance per frame of \name{} for individual actuations on the Jetson AGX Xavier}
    \label{fig:ptz-breakdown}
\end{figure}

To emphasize \name{}'s understanding of PTZ actuation and changing object sizes, we created three sequences out of the second sequence (SEQ\_02) from the original PANDA dataset where only one type of actuation (pan, tilt, or zoom) is applied at customized moments.
In \refFigure{ptz-breakdown}, we show a per-frame performance breakdown for several strategies on these sequences for a medium latency SLO of 3500ms.
Here, we compare \aliasnonconservative{} to Remix with the same SLO (\aliasremix{}), the EfficientDet-D6 variant as the best down-sampling option (\textbf{DS D6}), and uniform tiling with Efficientdet-D3 as the uniform choice that runs in roughly the same time (\textbf{\aliasuniform{} D3}).
Note that \name{} is always on top of the mAP curve for each frame, independently of the PTZ actuation used.
While the performance for the pan case exhibits less variance (object sizes remain rather constant), variations are more pronounced and sudden for the tilt and zoom cases.
For the tilt sequence (\refFigure{ptz-breakdown}, middle), we can see that when the camera tilts towards the bottom part of the global view and objects become larger as they approach the camera, the performance of Remix and the uniform approach substantially drops while the down-sample approach benefits from these larger object sizes.
Similar behavior can be seen for the zoom case (\refFigure{ptz-breakdown}, bottom).
\name{} correctly identifies these situations and rapidly adapts by either choosing plans closer to down-sampling or uniform tiling, whichever is beneficial in the situation.

\subsection{Ablation Studies}

In this section, we evaluate how the parameters and components of \name{} contribute to its performance. 

\subsubsection{Model Profiling}
\label{subsec:ablation-profile}
We first evaluate in \refFigure{eval-profile} how our non-linear, relative profile significantly improves the performance of adaptive strategies for the steerable case.
The figure shows the mAP achieved by \name{} in non-conservative mode using both our profile (UNCOP) and Remix's absolute size profile (UNCRP) as well as Remix using our profile (ROP) or theirs (RRP).
As it can be seen, both strategies benefit greatly from the use of our new profile.
Besides the improved performance of Remix with our profile, \name{} still provides an improvement in mAP between $0.98$-$1.40\times$ in comparison for the Xavier and $1.1$-$1.35\times$ for the Orin.
\name{} performs better for all SLOs 
while offering comparable mAP results for the others, indicating that the profile, while beneficial, is not the sole reason behind \name{}'s improved performance. 

\subsubsection{Tile Planning}

Most algorithm time is spent on tile planning, so we analyze it in detail.
We measured how often \name{} selects plans from the DP-DP algorithm or the uniform tiler.
\name{}'s non-conservative mode only chose DP-DP plans for SLOs below 6500ms.
As the SLOs increase, more frames select uniform plans.
On the Xavier, DP-DP plans were chosen for $76.64\%$ and $76.84\%$ of frames under 6500ms and 7000ms SLOs.
On the Orin, DP-DP plans were used exclusively for SLOs below 2750ms, and for $75.81\%$, $76.43\%$, $76.63\%$, and $76.84\%$ of the frames under 2750, 3000, 3250, and 3500ms SLOs, respectively.
This shows that uniform tiling supplements DP-DP when needed, while DP-DP still generates highly accurate plans within a limited search space.

\subsubsection{Object Extraction}
\label{subsec:ablation-historical-frames}

We evaluate how historical frames impact \name{}'s accuracy.
Experiments showed that varying the number of historical frames had little effect on mAP across different SLOs.
Using from 10\% to 30\% of frames results in no significant accuracy differences, indicating that \name{} does not need a high number of historical frames.
As long as the object distribution is well-represented, \name{} accurately predicts object locations and sizes.
Additionally, using uniform tiling with the best model to extract objects performs similarly to using ground truth data, therefore being a sufficient object extraction strategy.

\begin{table}
\caption{AP50 results achieved by \name{} and the Baselines for the different hours in our real PTZ data.}
\resizebox{\columnwidth}{!}{%
\begin{tabular}{|c|cc|cc|cc|}
\hline
\multicolumn{1}{|c|}{AP50} & \multirow{2}{*}{\begin{tabular}[c]{@{}c@{}}Down-smpling\\ D6\end{tabular}} & \multirow{2}{*}{\begin{tabular}[c]{@{}c@{}}Uniform\\ Partition D2\end{tabular}} & \multirow{2}{*}{\begin{tabular}[c]{@{}c@{}}Remix   \\ SLO 2000\end{tabular}} & \multicolumn{1}{c|}{\multirow{2}{*}{\begin{tabular}[c]{@{}c@{}}Uirapuru\\ SLO 2000\end{tabular}}} & \multicolumn{2}{c|}{\begin{tabular}[c]{@{}c@{}}Uirapuru's\\ Difference   to\end{tabular}} \\ \cline{1-1} \cline{6-7} 
\multicolumn{1}{|c|}{Hour} &                                                                            &                                                                                 &                                                                              & \multicolumn{1}{c|}{}                                                                             & Uniform                                      & Remix                                      \\ \hline
10:00 & 35.74\%             & 70.05\%            & 70.41\%  & \textbf{74.63\%}  & 4.58\%                                      & 4.22\%                                    \\
13:00 & 26.69\%             & 67.90\%            & 69.96\%  & \textbf{73.19\%}  & 5.29\%                                      & 3.22\%                                    \\
16:00 & 28.62\%             & 72.12\%     & 70.87\%        & \textbf{77.14\%}  & 5.02\%                                      & 6.27\%                                    \\
19:00 & 25.84\%             & 63.29\%            & 65.89\% & \textbf{68.88\%}  & 5.59\%                                      & 2.99\%                                    \\ \hline
\end{tabular}
}
\label{tab:case_study_ap50}
\end{table}

\subsection{Case Study}
\label{sec:case_study_intro}

In addition to experiments with our created steerable PANDA dataset, we also evaluate \name{}
in a real-world setup. 
We collected 4K video sequences (around 6,400 frames in total) from a steerable camera in a square at different hours of the day\footnote{The use of the data has been approved by our university ethics committee.}. 
The sequences were collected on a sunny day at 10:00, 13:00, 16:00, and 19:00, covering different lighting conditions and exhibiting average person densities of 73, 211, 233, and 147, respectively.

The steerable camera performs several PTZ actuations in a fixed loop.
The actuation loop is pre-defined by the camera operator, without any interference from us, and it was always exactly the same throughout the different times of the day.
The high degree of Pan and Tilt causes objects to occlude others constantly.
When Zoom is applied, the camera executes abrupt actuation, but it is limited in duration (less than 7\% of the loop sequence) and magnification (only a 2× zoom factor).
Each sequence consists of three consecutive actuation loops.  
We use the first loop as our historical frames for each sequence and the other two for evaluation.
Similarly to Remix~\cite{2021-mobicom-remix}, we generate labels using the best possible strategy as our oracle model (Uniform Partition with EfficientDet D6) due to the lack of ground-truth labels for the collected sequences.

\subsubsection{Overall Performance}
\label{subsec:case_study}

For this analysis, we set the latency SLO to 2,000ms. We compared \name{} against Remix, the best down-sampling approach (EfficientDet D6), and Uniform Partition using EfficientDet D2 since it has the execution time closest to the SLO.
Like Remix~\cite{2021-mobicom-remix}, we use the AP50 in our case study.
However, complete results for the mAP metric, which confirm the observed trends, can be found in Appendix~\ref{appendix:case_study_metrics}.

\refTable{case_study_ap50} shows a comparison against the baselines for different times of the day.
\name{} achieves the highest AP50 at all times.
Remix and Uniform Partition have similar results, but Uniform Partition performs better at 16:00.
As detailed in \refSubsection{ptz_breakdown}, Remix becomes suboptimal under strong Pan and Tilt motion: plans that should prioritize dense or complex areas are instead executed in easier or empty regions.
Frequent camera motion also makes frame skipping harder, further limiting Remix’s ability to adapt.
When objects are more evenly distributed, Remix can perform worse than Uniform Partition, as observed at 16:00. 
In contrast, \name{} can address both issues with its per-frame plan creation and improve object detection in the scene.
Additionally, larger Zooms can drastically degrade the performance of Remix and Uniform while \name{} remains robust, widening the performance gap. 
We can see that \name{} offers tiling solutions that can cope with those changes and maintain higher mAP performance.
However, Zoom duration and magnification were limited by the operator in our case study, so these effects are less prominent than they would be in scenarios that require closer inspection.

\begin{table}[]
\caption{AP50 values of \name{} when using the plans generated with the historical frames of different times.}
\small
\begin{tabular}{|c|cccc|}
\hline
AP50          & \multicolumn{4}{c|}{Update at}                                                                                                    \\ \hline
Evaluation at & \multicolumn{1}{c}{10:00} & \multicolumn{1}{c}{13:00} & 16:00                             & 19:00                     \\ \hline
10:00      & \textbf{74.63\%}             & \multicolumn{1}{c}{-}        & -                                    & -                            \\
13:00      & \textbf{73.91\%}             & 73.19\%                      & -                                    & -                            \\
16:00      & 77.09\%                      & 76.30\%                      & \multicolumn{1}{l}{\textbf{77.14\%}} & -                            \\
19:00      & \textbf{69.81\%}             & 68.26\%                      & \multicolumn{1}{l}{68.93\%}          & \multicolumn{1}{l|}{68.88\%} \\ \hline
\end{tabular}
\label{tab:case_study_update}
\end{table}

\subsubsection{Performance Under Different Update Rates}
\label{subsubsection:update_performance}

As the object distribution changes over time, \name{} may need to update its object history by collecting new frames.

\refTable{case_study_update} shows how historical frames collected at different times might affect \name{}.
The results show that staleness in historical frames has minimal impact, with performance differences under 1\%.
As long as object distribution remains similar, \name{} does not need to re-run its bootstrap phase. 
\name{}'s runtime estimator (in~\refSubsection{plan-runtime-estimation}) calculates the inference time SLO using the model profiles and the object history.
At 10:00, the number of objects is smaller, creating less overhead in the system. 
Therefore, the historical frames from 10:00 yield the best performance, even 9 hours later.
This overhead will be discussed in~\refSubsection{inference_overhead}.

\subsubsection{Performance Under Different Weather Conditions}
\label{subsec:case_study_weather}

The collected 4K videos also exhibit different lighting and weather conditions (namely sunny, cloudy, and dark) (see in Appendix~\ref{appendix:dataset_figures}).
The sunny and cloudy sequences were recorded at 13:00, while the dark sequence was captured at 19:00.

\refTable{case_study_weather} shows that Uirapuru performs best, independent of these conditions.
We can also see that the cloudy condition gives better results than the sunny and dark ones for all approaches. 
This is mainly due to even lighting under the cloudy condition, eliminating shadows, flares, and brightness changes that hinder object detection.
When it is dark at night, artificial lighting intensifies these issues, further reducing accuracy.

\begin{table}[]
\caption{AP50 values achieved by \name{} under different weather conditions.}
\small
\begin{tabular}{|c|cccc|}
\hline
AP50         & \name{}         & Remix            & Uniform          & Downscale       \\ \hline
Sunny      & \textbf{73.19\%} & 69.96\%          & 67.90\%          & 26.69\%         \\
Dark    & \textbf{68.88\%} & 65.89\%          & 63.29\%          & 25.84\%         \\
Cloudy    & \textbf{75.46\%} & 71.91\%          & 71.33\%          & 29.11\%         \\ \hline
\end{tabular}
\label{tab:case_study_weather}
\end{table}

\begin{figure}[t]
    \centering
    \includegraphics[width=\twocolgrid,page=1]{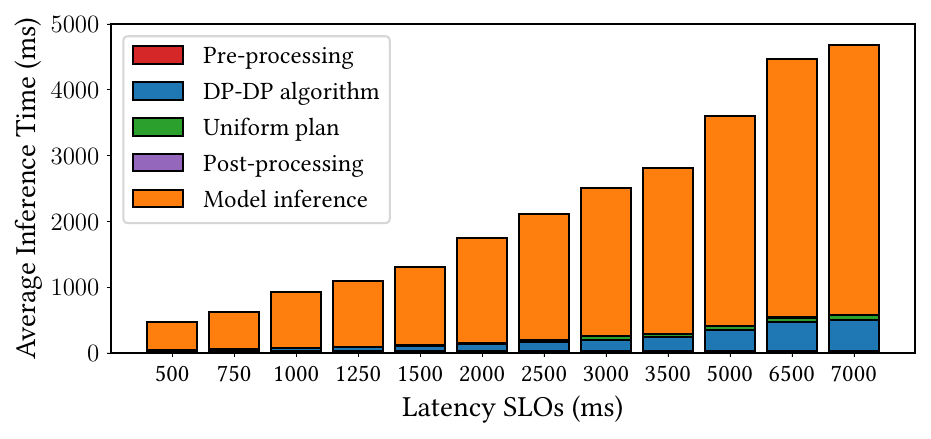}
    \caption{Per-frame break-down of the time spent on pre-processing, tile planning, post-processing, and model inference  across different latency SLOs.}
    \Description{Per-frame break-down of the time spent on pre-processing, tile planning, post-processing, and model inference}
    \label{fig:eval-timings}
\end{figure}

\subsection{Overhead Analysis}
\label{subsec:inference_overhead}

Next, we analyze the extra overhead introduced by \name{}.
\subsubsection{Execution Time}

\refFigure{eval-timings} shows the average execution time of each step on the Jetson Xavier. 
Pre-processing, which consists of extracting and transforming historical objects, computing the object distribution for each quad-tree node, and estimating the expected accuracy for each node-model pair, introduces minimal overhead, consuming only 5.14\% of the available SLO at 500ms and 0.51\% at 7000ms.
Post-processing,which merges detections using Non-Maximum Suppression (NMS), also has low overhead, ranging from 0.59\% at 750ms to 0.15\% at 7000ms.
The uniform plan computation is fast, contributing 0.24-2.02\% of the SLO, while DP-DP introduces the largest overhead, from 4.37\% at 500ms to 10.22\% at 7000ms. 
Generally, \name{}'s overhead remains low (less than 12.5\% for all SLOs), with most time spent on model inference (87\% on Xavier, 89\% on Orin).

In our case study in~\refSubsection{case_study}, the pre-processing stage can be affected by the number of objects in the historical frames.
As object numbers grow, extraction, transformation, and distribution computations take longer, reducing the inference time and affecting accuracy. 
Furthermore, \name{} generates a new plan every frame, even when actuation is minimal, leading to redundant computations.

\subsubsection{Memory Footprint}
\label{subsection:memory_footprint}

\name{} can process each frame using multiple models.
Thus, all models (from D0 to D6) must be loaded and initialized into memory during the bootstrap phase.
Our measurements show that \name{} consumes a maximum of
12.2 GB memory during inference on both Jetsons. 
Approximately half of this memory is used by D5 and D6, which consume nearly 6.1 GB of memory combined.

\subsubsection{Energy Comsumption}
\label{subsection:energy_footprint}

\begin{figure}[t]
    \centering
    \includegraphics[width=\twocolgrid,page=1]{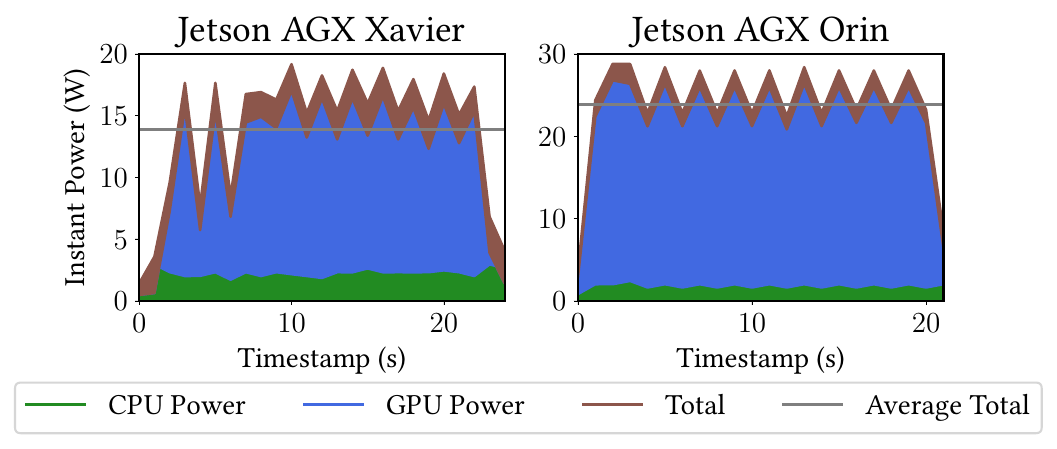}
    \caption{Energy consumption trace when processing ten
frames with Jetson Xavier (left) and Orin (right).}
    \Description{Power consumption levels of \name{} on the Jetson Xavier on the left and on the Jetson Orin on the right}
    \label{fig:eval-power}
\end{figure}

\refFigure{eval-power} shows the power consumption of \name{} on both hardware. 
We used Jetson's built-in \texttt{tegrastats} monitor to collect the instant energy consumption traces. 
We executed 10 frames using an SLO of 2000 ms. The total power consumption per frame is, on average, about 13.9 W for the Jetson Xavier and 23.8 W for the Orin.
The GPU consumes most of the power in both, with an average of 11.7 W in the Xavier and 22 W in the Orin. 
The power consumption of \name{} matches those achieved by Remix~\cite{2021-mobicom-remix}, and the total average consumption represents less than half of the nominal power of the devices.

\section{Discussion and Limitations}
\label{sec:discussion}

In the following, we'd like to discuss the limitations of our approach and possible opportunities for further exploration.

\myparagraph{Steerable Camera Parameters}
\name{} assumes that the steerable camera is mounted in a fixed position, as PTZ cameras are.
This is necessary for the distribution controller to process the object history and create the current-view object distribution.
For example, video sequences from unmanned aerial vehicles (drones) that are not mounted in a fixed position would imply more complex actuation. 
This would require a much more robust and time-consuming version of the distribution controller.

\myparagraph{Historical Objects}
Selecting historical frames for high-quality partition plans remains an open question.
Extensive object histories increase pre-processing overhead.
One solution is to reduce the sampling rate when collecting objects.
Another solution would be to group objects in regions and store pre-processed distributions.
While this would reduce the overall overhead linked to the number of objects in the object history, it would also introduce a new challenge: deciding how to merge these objects.
Therefore, we leave these directions for future work.

\myparagraph{Reducing Runtime Overhead}
\name{} uses fixed parameters in the DP-DP algorithm, which could be optimized to reduce the overhead as the latency budget grows.
Dynamically adjusting parameters, such as increasing the latency interval (step size) to shrink the inference latency budget array, could lower the runtime without impacting accuracy. 
Additionally, skipping tile plan creation for frames with minimal or no actuation could reduce overhead, allowing better plans to be used when updates are unnecessary.
Finally, more effort is required to optimize memory/energy usage.

\myparagraph{Memory Bottleneck}
\name{}’s performance is influenced by a trade-off between memory capacity and the available model pool.
Larger models provide higher detection accuracy, while smaller ones enable faster inference.
A diverse pool of models would enrich the optimization space of \name{}, leading to better performance.
However,  loading more models into memory is not always possible and might exceed the capacity of many edge platforms (e.g., Jetson Nano).  
To address this, \name{} can rely on a smaller pool of models or utilize a more optimized framework for deploying machine learning models on resource-constrained devices like LiteRT~\cite{2025-website-litert}.
Such reductions come at the cost of accuracy, even though reducing the pool of models equally affects baseline accuracy.
While we use EfficientDet as an example throughout this paper, \name{} is agnostic to the underlying models and can operate with any compatible neural networks.

\myparagraph{Skipping}
Currently, tile skipping is only done implicitly in our DP-DP algorithm, but not explicitly during the inference in the runtime phase  (as done e.g., in Remix where tiles are marked as empty if they were empty in previous frames).
Skipping on a steerable camera scenario is less trivial than in the static case; actuation can quickly change object distributions between frames. 
However, more latency budget could be gained when performed effectively.
Developing a skipping mechanism that works in the steerable context could be a promising avenue for future work.


\section{Conclusion}
\label{sec:conclusion}

We introduce \name{}, a framework for latency-constrained object detection for \textbf{high-resolution steerable} cameras.
\name{} tackles the challenges introduced by persistent camera actuation and changing viewpoints by incorporating an understanding of that movement into its system design. 
Additionally, \name{} proposes a new efficient way of generating adaptive tile plans quickly for each video frame.
Our evaluation with synthetic PTZ video datasets shows that \name{} improves accuracy by up to $1.45\times$ for the same inference time and produces results with similar accuracy with a $4.53\times$ inference speedup compared to previously existing approaches.
Our experiments with real-world videos collected from an actual PTZ camera further confirm the significant improvements brought by \name{}.


\begin{acks}
This work is part of the  Real-Time Video Surveillance Search project (grant number 18038), financed by the Dutch Research Council (NWO). 
\end{acks}

\bibliographystyle{ACM-Reference-Format}

\newpage
\appendix
\section*{Appendix}

\section{Pseudo Code}
\label{appendix:pseudo}

\refAlgorithm{dp-dp} presents the recursive pseudo-code for the DP-DP algorithm, initiated with the call \texttt{DP-DP(root\_node, $\mathbf{0}$, $\mathbf{0}$)}. 
The algorithm determines the optimal model assignment for each node across all discrete latency values (lines $9$-$15$). 
These latency values range from zero to the maximum latency budget (i.e., the knapsack capacity) in steps defined by a latency interval.
The algorithm then starts recursively constructing the partial and optimal DP solution at each node in a depth-first post-order manner (lines $6$-$8$).
At each parent node, the algorithm compares the local optimal solution with the children's optimal solution to choose the best one (lines $19$-$21$), satisfying constraint \textbf{C1}. 
For each node and latency value, there are $|\mathbb{M}|$ model choices, and the algorithm selects only the best model, satisfying constraint \textbf{C2} (lines $10$-$15$).
Note that all dynamic programming algorithms depend on previous partial and optimal solutions.
In our algorithm, the previous solution may come from the left sibling or the left sub-tree.
If a node lacks a left sibling or left sub-tree, we use an array of zeros.
Additionally, while traversing, we keep track of which models are assigned to which nodes.
The algorithm ultimately returns an array with the optimal mapping of nodes to models, where every node is either assigned a model or marked as empty.

\begin{algorithm2e}[tb]
\DontPrintSemicolon
\IncMargin{0.25em}
\SetNoFillComment

\definecolor{FunctionNameColor}{RGB}{153, 0, 0}
\colorlet{CommentColor}{green!70!black!180}
\newcommand\commentfont[1]{\scriptsize\ttfamily\textcolor{CommentColor}{#1}}
\SetCommentSty{commentfont}

\newcommand{\tcccomment}[1]{\tcp{#1}}

\SetAlCapNameFnt{\small}
\SetAlCapFnt{\small}

\footnotesize


\SetKwProg{Fn}{Function}{:}{}

\SetKwFunction{FMain}{\textcolor{FunctionNameColor}{DP-DP}}
\SetKwFunction{GetChildren}{CHILDREN}
\SetKw{And}{\textbf{and}}
\SetKwData{nodeID}{$n$}
\SetKwData{prevDPRow}{$DP_{left}$}
\SetKwData{prevMappingRow}{${SOL}_{left}$}
\SetKwData{latBudget}{$L_{slo}$}
\SetKwData{modelAccuracy}{$A$}
\SetKwData{modelLatency}{$L$}
\SetKwData{dpRow}{$DP$}
\SetKwData{mappingRow}{$SOL$}
\SetKwData{childID}{$c$}
\SetKwData{children}{$C$}
\SetKwData{zeros}{$0$}
\SetKwData{setOfNodes}{$\mathbb{N}$}
\SetKwData{numNodes}{$|\mathbb{N}|$}
\SetKwData{tempDPRow}{$DP_{temp}$}
\SetKwData{tempMappingRow}{${SOL}_{temp}$}
\SetKwData{dpChildRow}{$DP_{child}$}
\SetKwData{mappingChildRow}{${SOL}_{child}$}
\SetKwData{colIter}{$i$}
\SetKwData{modelIter}{$m$}
\SetKwData{setOfModels}{$\mathbb{M}$}
\SetKwData{numModels}{$|\mathbb{M}|$}
\SetKwData{prevJ}{$j$}
\SetKwData{bestAccuracy}{$a_{best}$}
\SetKwData{bestAccuracy}{$\dpRow(\colIter)$}
\SetKwData{accuracy}{\modelAccuracy(\nodeID, \modelIter) + \prevDPRow(\prevJ)}
\SetKwData{lastChildDPRow}{$DP_{lastchild}$}
\SetKwData{lastChildMappingDPRow}{$SOL_{lastchild}$}


\KwIn{A node $\nodeID$ with optimal accuracy values $\prevDPRow$ and mapping solution $\prevMappingRow$ from the left subtree or sibling}
\KwOut{$\dpRow$ array with optimal accuracy values and model mapping solution $\mappingRow$ for every latency value at node $\nodeID$}
\KwData{Matrix $\modelAccuracy$ representing node accuracy per model, array $\modelLatency$ storing latency values for each model, latency SLO $\latBudget$, set $\setOfNodes$ of nodes, and set $\setOfModels$ of models}  

\BlankLine
\Fn{\FMain{\nodeID, \prevDPRow, \prevMappingRow}}{

    $\dpRow(i) \gets \zeros, \forall i \in [0, \latBudget]$\;
    $\mappingRow(i, j) \gets \zeros, \forall i \in [0, \latBudget], \forall j \in [0, \numNodes - 1]$\;

    \tcccomment{Traverse all the children from left to right}
    $\children \gets \GetChildren(\nodeID)$\;
    $\tempDPRow \gets \prevDPRow, \tempMappingRow \gets \prevMappingRow$\;

    \ForEach{$\childID \in \children$}{
        $\dpChildRow, \mappingChildRow \gets \FMain(\childID, \tempDPRow, \tempMappingRow)$\;
        
        $\tempDPRow \gets \dpChildRow, \tempMappingRow \gets \mappingChildRow$\;
    }

    \tcccomment{Compute DP array values (accuracy) for this node}
    \For{$\colIter \gets 1$ \KwTo $\latBudget$}{
        \For(\tcp*[f]{Select the best model}){$\modelIter \gets 0$ \KwTo $\numModels - 1$}{
            $\prevJ \gets \colIter - \modelLatency(\modelIter)$\;        
            \If{$\prevJ \geq 0$ \And ($\accuracy) > \bestAccuracy$}{
                $\bestAccuracy \gets \accuracy$\;
                $\mappingRow(\colIter, k) \gets \prevMappingRow(\prevJ, k),  \forall k \in [0, \numNodes - 1]$\;
                $\mappingRow(\colIter, \nodeID) \gets \modelIter + 1$\;
            }
        }

        \tcccomment{Select the left subtree/sibling solution if better}
        \If{$\prevDPRow(\colIter) > \bestAccuracy$}{
            $\bestAccuracy \gets \prevDPRow(\colIter)$\;
            $\mappingRow(\colIter, k) \gets \prevMappingRow(\colIter, k), \forall k \in [0, \numNodes - 1]$\;
        }

        \tcccomment{Select the children solution if better}
        \If{$\children \neq \emptyset$ \And $\lastChildDPRow(\colIter) > \bestAccuracy$}{
            $\bestAccuracy \gets \lastChildDPRow(\colIter)$\;
            $\mappingRow(\colIter, k) \gets \lastChildMappingDPRow(\colIter, k), \forall k \in [0, \numNodes - 1]$\;
        }
    }
    
    \Return{$\dpRow$, $\mappingRow$}\;
}

\caption{Tiles Selection and Model Mapping}
\label{alg:dp-dp}
\end{algorithm2e}

\section{Optimality Proof}
\label{appendix:proof}
Dynamic programming framework is based on the \emph{principle of optimality}~\cite{1954-bellman-dptheory}. It divides the problem into smaller sub-problems, finds their optimal solutions, and iteratively combines these to form the optimal solution to the original problem.

\subsection{Recurrence Relation} In our context, we iterate over the tree (representing image space partitioning) to compute the list of nodes and their corresponding models from the model set $\mathrm{M}$ that together achieve the optimal value (aggregate estimated accuracy) for a given latency budget $L_{slo}$. We construct the value (aggregate estimated accuracy) matrix $DP(n, l)$ at each node $n$ for discrete latency budgets $l$ by traversing the tree in a depth-first post-order manner. 
The \emph{recurrence} relation is defined as follows:
\begin{small}
    \begin{subequations}
        \begin{align}
            DP(n, l) &= \max_{m \in \mathrm{M} \text{ and } l \ge l_m} DP(n_{left}, l - l_m) + A(n, m) \label{eq:proof:recurrence:step1} \\
            DP(n, l) &= \max \{DP(n, l), DP(n_{left}, l), DP(n_{rchild}, l)\} \label{eq:proof:recurrence:step2}
        \end{align}
    \end{subequations}
\end{small}

Here, $A(n, m)$ represents the estimated accuracy when model $m$ is selected for node (or tile) $n$, and $l_m$ denotes the inference latency of model $m$. In Equation~\ref{eq:proof:recurrence:step1}, the recurrence function first computes $DP(n, l)$ based on the previous sub-solution at node $n_{left}$, which is the nearest left sibling node on the ancestor path from node $n$ to the root. The aim is to select the model that gives the best accuracy. Then, in Equation~\ref{eq:proof:recurrence:step2}, we exclusively select the optimal sub-solution from the current sub-solution, the previous sub-solution, and the sub-solution of the rightmost child node (satisfying Constraint \textbf{C1}, see \S\ref{subsec:tile-planner}).

\subsection{Proof of Correctness}
We now consider the cases one by one for different types of nodes in the tree and then use mathematical induction on the tree structure to establish the correctness of the algorithm. 

\myparagraph{Base Case} Since the root node does not have a nearest left sibling node on its ancestor path, we set $DP(n_{left}, l)$ at the root node to zero for latency budget $l$. This means $DP(n_{left}, l)$ will be zero for all nodes along the leftmost path. Similarly, $DP(n_{rchild}, l)$ will be zero for all leaf nodes for any latency budget $l$, as leaf nodes have no children.

\myparagraph{Case 1: Leftmost Leaf Node} This node is the first visited node in our depth-first post-order traversal. It has no $n_{left}$ or $n_{rchild}$. Following Equation~\ref{eq:proof:recurrence:step1} and the base case, the DP value for this node simply depends on the model that maximizes the accuracy under latency budget $l$, given by $DP(n, l) = \max_{m \in \mathrm{M} \text{ and } l \ge l_m} A(n, m)$. Since there are no other nodes to consider, the DP value at this node is correctly initialized to the best achievable accuracy within the latency budget.

\myparagraph{Case 2: Right Sibling of the Above Node} This is the second node visited. It has $n_{left}$ (the above node) but no $n_{rchild}$. Equation~\ref{eq:proof:recurrence:step1} ensures that the best possible model $m$ is considered in combination with the optimal solution from its left sibling $n_{left}$  for a given latency budget $l$. Equation~\ref{eq:proof:recurrence:step2} helps the algorithm to explore whether including this node with a specific model
 gives a better solution than considering $n_{left}$ alone. This guarantees optimality by evaluating all possible combinations between the two nodes.

\myparagraph{Case 3: Parent of the Above Two Nodes} For simplicity and without loss of generality, we assume that the tree is binary. Hence, this node is the third node visited. It has no $n_{left}$ but has $n_{rchild}$. Equation~\ref{eq:proof:recurrence:step1} computes the DP value by choosing the model that maximizes accuracy under the latency budget $l$, given by $DP(n, l) = \max_{m \in \mathrm{M} \text{ and } l \ge l_m} A(n, m)$. Since this node has children, Equation~\ref{eq:proof:recurrence:step2} ensures that the final optimal DP value results from either considering this node's contribution or a better solution from the combination of child nodes stored at the rightmost child, $n_{rchild}$.  

\myparagraph{Case 4: Leaf Nodes without Immediate Left Sibling} These nodes lack an immediate left sibling but have a nearest left sibling on the ancestor path. Equation~\ref{eq:proof:recurrence:step1} ensures that we consider the contribution of this node $n$ with the best model in combination with the optimal solution at $n_{left}$. Equation~\ref{eq:proof:recurrence:step2} ensures the final optimal DP value for node $n$ is the maximum of either including the contribution from node $n$ or only the optimal solution from $n_{left}$. This case is similar to Case $1$.

\myparagraph{Case 5: Nodes with Left Sibling and Children} This is the most general case. These nodes have both a left sibling and children. Equation~\ref{eq:proof:recurrence:step1} ensures that we select the best model $m$ in combination with the optimal solution at $n_{left}$. Equation~\ref{eq:proof:recurrence:step2} ensures the DP value is optimal by accounting for all contributions from its left sibling, its children (the solution that is built on top of the left sibling solution), and the node itself. 

Our depth-first post-order traversal ensures that for any node $n$, the optimal solution at its rightmost child considers the combination of all the descendant nodes that is included on top of the optimal solution at the left sibling. This approach guarantees that choosing the best solution between the children (descendants) and the node itself satisfies Constraint \textbf{C1}.

We now present the proof of correctness by induction. Let $n \in [0, N]$ represent the index in the ordered sequence of tree traversal, where $n=0$ corresponds to the case of no nodes. We define $(n, l) < (n', l')$ if $n < n'$ and $l \le l'$.

\myparagraph{Induction Hypothesis} The algorithm computes the value of $DP(n, l)$ correctly for all $(n, l) < (n', l')$, meaning all node indices from $0$ to $n$ are correctly processed.

\myparagraph{Base Case}
$DP(n, l) = 0$ for $n = 0$ and $\forall l \in [0, L_{slo}]$. This is trivially correct since there are no nodes. Also, $DP(n, l) = 0$ for all $n$ and $l < \min_{m \in \mathrm{M}} l_m$ because we cannot select any model without violating the latency constraint.

\myparagraph{Induction Step} To compute $DP(n', l')$, we need the values of $DP(i, l')$, $DP(i, l' - l_m)$ for all $m \in \mathrm{M}$, and $DP(j, l')$ as per the recurrence relation. Here, $i$ represents the index of the nearest left sibling node on the ancestor path, where $i = n' - 1$ if the node lacks children, and $j$ represents the index of the rightmost child, where $j = n' - 1$ if the node has children, otherwise $0$. The depth-first post-order traversal guarantees both $i < n'$ and $j < n'$. Thus, by the inductive hypothesis, the values $DP(i, l')$, $DP(i, l' - l_m)$ for all $m \in \mathrm{M}$, and $DP(j, l')$ are available and computed \emph{correctly}.

The algorithm determines the optimal value for $DP(n', l')$ by comparing three options: using $DP(i, l' - l_m)$ if adding this node to the solution list is beneficial, using $DP(j, l')$ if the combination of child nodes is better, or retaining the previous solution $DP(i, l')$ if adding any node in the subtree rooted at this node is not beneficial. Therefore, the value for $DP(n', l')$ is computed \emph{correctly}.

Overall, our DP-DP algorithm provides an optimal solution for the following reasons: (1) Principle of Optimality~\cite{1954-bellman-dptheory}: The solution is constructed by combining solutions to subproblems, whose correctness is proven above. (2) Exhaustive Search in Subproblems: The recurrence relation considers all possible models $m$ at each node $n$ and all possible ways to partition the latency budget between the node and its subproblems. (3) No Overlapping Subproblems: Each subproblem is computed exactly once and does not incur redundant computations.

\section{Case Study mAP Metrics}
\label{appendix:case_study_metrics}

Here are the results of additional metrics for our case study analysis. \refTable{case_study_mAP} shows the mAP results achieved by Uirapuru and Remix using a budget of 2000 ms; Uniform Partition using EfficientDet-D2, which is the closest Uniform Partition network approach to the budget in terms of execution time; and down-sampling using Efficiendet D6, which is the best down-sampling approach available.

\refTable{case_study_update_map} shows the mAP results reached by \name{} when the object history is updated at different moments than the evaluation and the evaluation moment itself. As it can be seen, \name{} reaches the highest mAP results while using the object history generated at 10:00. These results are on par with the ones seen in \ref{subsubsection:update_performance}. The only difference happens at 16:00, where the result using the object history from 10:00 is now higher than the one using the history of 16:00. However, the difference is only $0.03\%$.

\begin{table}[!t]
\caption{mAP results achieved by Uirapuru and the Baselines for the different hours in our real PTZ data.}
\label{tab:case_study_mAP}
\resizebox{\columnwidth}{!}{%
\begin{tabular}{|c|cc|cc|lc|}
\hline
mAP     & \multicolumn{2}{c|}{Corresponding  Best} & \multicolumn{2}{c|}{Budget   2000} & \multicolumn{2}{c|}{\begin{tabular}[c]{@{}c@{}}Uirapuru's\\ Difference to\end{tabular}} \\ \hline
Hour     & Down-sampling       & Uniform            & Remix          & Uirapuru          & \multicolumn{1}{c}{Uniform}                           & Remix                           \\
10:00:00 & 18.54\%             & 41.45\%            & {\ul 47.34\%}  & \textbf{50.17\%}  & 8.72\%                                                & 2.83\%                          \\
13:00:00 & 12.10\%             & 38.32\%            & {\ul 43.46\%}  & \textbf{45.42\%}  & 7.10\%                                                & 1.96\%                          \\
16:00:00 & 13.33\%             & {\ul 42.14\%}      & {\ul 45.33\%}  & \textbf{49.00\%}  & 6.86\%                                                & 3.67\%                          \\
19:00:00 & 12.28\%             & 35.08\%            & {\ul 40.67\%}  & \textbf{42.30\%}  & 7.22\%                                                & 1.64\%                          \\ \hline
\end{tabular}%
}
\end{table}

\begin{table}[!t]
\caption{mAP values of Uirapuru when using the plans generated with the historical frames of different times.}
\label{tab:case_study_update_map}
\begin{tabular}{|c|cccc|}
\hline
mAP           & \multicolumn{4}{c|}{Update at}                            \\ \hline
Evaluation at & 10:00         & 13:00 & 16:00         & 19:00 \\ \hline
10:00      & \textbf{50.17\%} & -        & -                & -        \\
13:00      & \textbf{45.97\%} & 45.42\%  & -                & -        \\
16:00      & \textbf{49.03\%} & 48.77\%  & 49.00\% & -        \\
19:00      & \textbf{42.81\%} & 42.31\%  & 42.28\%          & 42.30\%  \\ \hline
\end{tabular}
\end{table}

\refTable{case_study_weather_map} shows the mAP values achieved by \name{}, Remix, down-sampling, and Uniform Partition under different lighting conditions. The trends here are the same as in \ref{subsec:case_study_weather}.

\begin{table}[!t]
\caption{mAP values achieved by Uirapuru under different lighting conditions.}
\label{tab:case_study_weather_map}
\begin{tabular}{|c|lll|}
\hline
mAP      & \multicolumn{1}{c}{Sunny} & \multicolumn{1}{c}{Dark} & \multicolumn{1}{c|}{Cloudy} \\ \hline
Uirapuru  & \textbf{45.42\%}        & \textbf{42.30\%}          & \textbf{47.52\%}           \\
Remix     & 43.46\%                 & 40.67\%                   & 45.62\%                    \\
Uniform   & 38.32\%                 & 35.08\%                   & 41.72\%                    \\
Downscale & 12.10\%                 & 12.28\%                   & 12.78\%                    \\ \hline
\end{tabular}
\end{table}

\section{Dataset Inspection}
\label{appendix:dataset_figures}

\begin{figure*}[!t]
    \centering
    \includegraphics[width=0.95\textwidth]{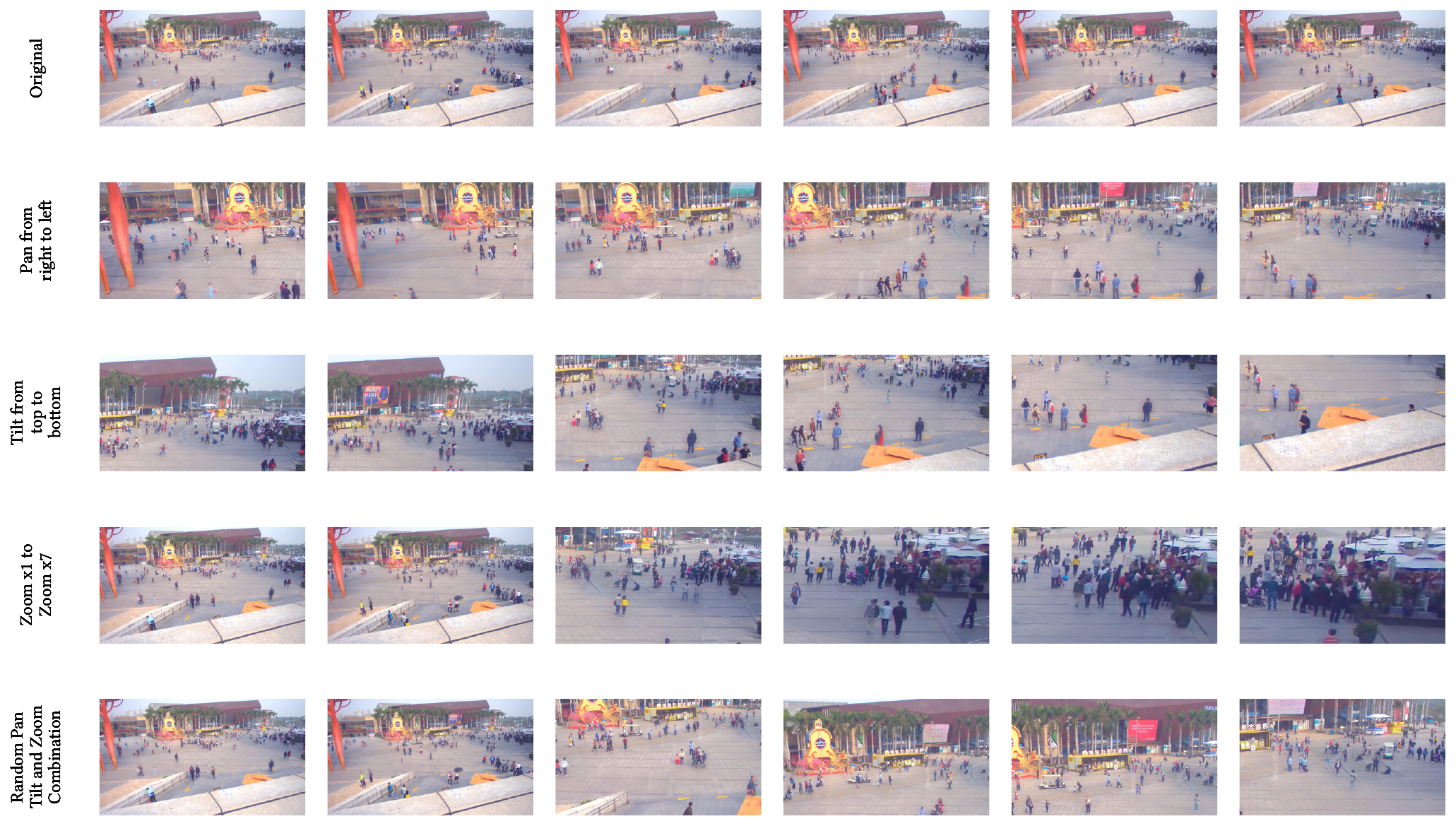}
    \caption{Original Panda sequence frames (top row) and its generated sequences examples for Pan (second row), Tilt (third row), Zoom (third row), and a random actuation combining Pan, Tilt and Zoom (bottom row).}
    \Description{Panda dataset figure examples}
    \label{fig:panda_examples}
\end{figure*}

Our scenario requires a dataset of high-resolution video frames with a high density of objects generated by a steerable camera, ground truth labels, and pre-defined/controlled camera movements.
Datasets in the literature lack one or more of those requirements.
The PANDA dataset~\cite{2020-cvpr-panda} is the one that gets closest to fulfilling all requirements, but it lacks pre-defined/controlled camera movements like all others.
We created a video dataset with movements based on the PANDA dataset~\cite{2020-cvpr-panda}. 
\refFigure{panda_examples} shows how we generated new frames that simulate Pan, Tilt, Zoom actuation, or a combination of all three.
Note that the camera transformation consists of rotation and scaling (zoom) and does not contain translation (we assume the camera is mounted in a fixed position).

\begin{figure*}[!t]
    \centering
    \includegraphics[width=0.95\textwidth]{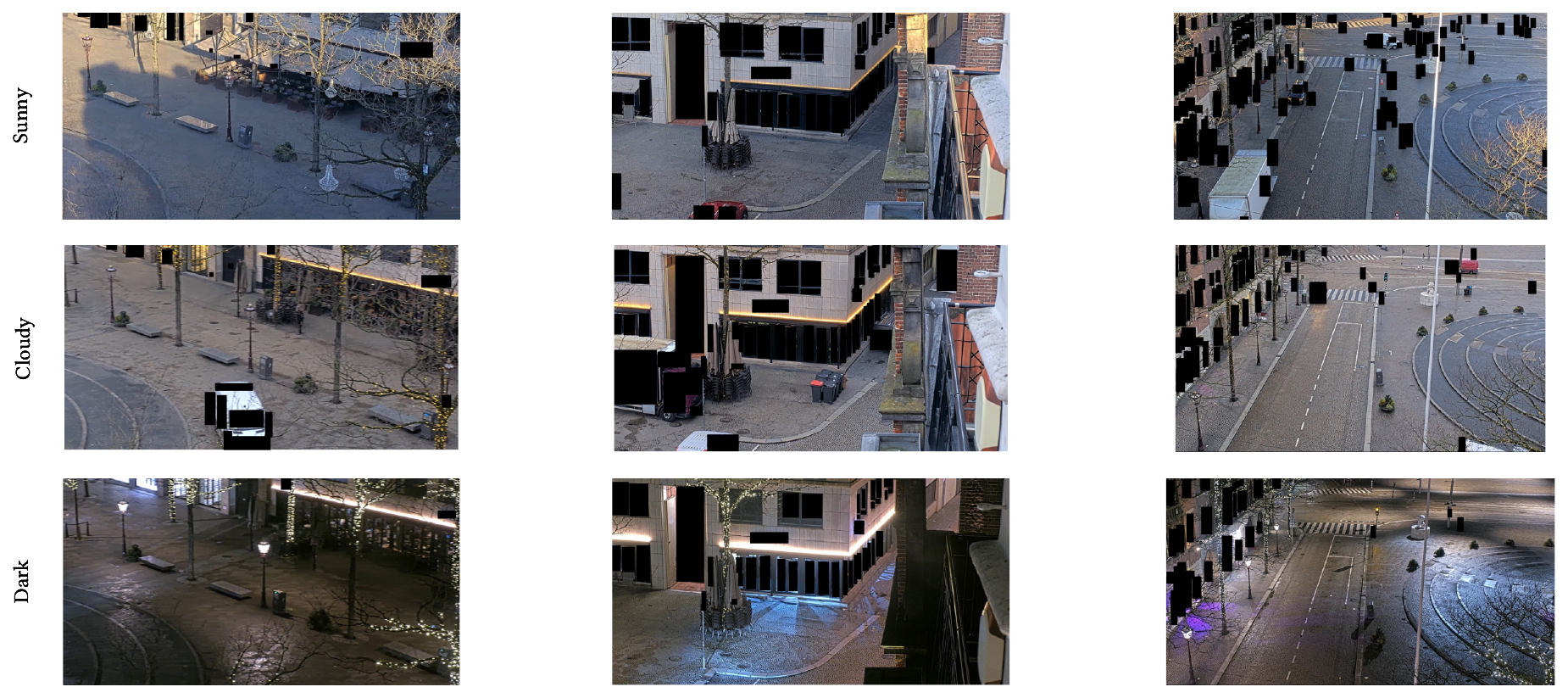}
    \caption{Case study frame examples at different weather/lighting conditions. Top row show frames captured during a sunny day, middle row shows frames captured at a cloudy day, and bottom row shows frames captured at night when it is dark.}
    \Description{Case study dataset figure examples}
    \label{fig:case_study_examples}
\end{figure*}

We also conducted a case study using a real steerable camera. 
The camera is in a city square, and
\refFigure{case_study_examples} shows some of the frames captured by steerable cameras in different weather/light conditions.
Each frame in the row was captured with a different actuation.
Persons, cars, windows, ads, and other objects are redacted for privacy preservation. 

\end{document}